\documentclass[times,twocolumn,final]{elsarticle}

\usepackage{lineno,hyperref}

\journal{Journal of \LaTeX\ Templates}

\usepackage{times}
\usepackage{epsfig}
\usepackage{amssymb}
\usepackage{bbm}
\usepackage{xcolor}
\usepackage{amsmath} 
\usepackage{multirow,booktabs}
\usepackage{mathtools}
\usepackage{soul}
\usepackage{xcolor}         
\usepackage{bm}             
\usepackage[capitalise]{cleveref}       
\usepackage{graphicx}       
\usepackage{booktabs}       
\usepackage{amsfonts}       
\usepackage{nicefrac}       
\usepackage{microtype}      

\mathchardef\mhyphen="2D 
  
\definecolor{red}{rgb}{0.234, 0.6835, 0.808}
\usepackage{color, colortbl}
\definecolor{Gray}{gray}{0.9}

\renewcommand{\hl}[1]{#1}

\usepackage{multirow}
\usepackage{tabularx}

\usepackage{algorithm}
\usepackage{algorithmicx}
\usepackage{etoolbox}
\usepackage{algpseudocode}

\algnewcommand\algorithmicinput{\textbf{Input:}}
\algnewcommand\INPUT{\item[\algorithmicinput]}
\algnewcommand\algorithmicoutput{\textbf{Output:}}
\algnewcommand\OUTPUT{\item[\algorithmicoutput]}

\newcommand{\algstrut}[1][\algruledefaultfactor]{\vrule width 0pt
depth .25\baselineskip height #1\baselineskip\relax}
\newcommand*{\algrule}[1][\algorithmicindent]{\hspace*{.5em}\vrule\algstrut
\hspace*{\dimexpr#1-.5em}}
\makeatletter
\newcount\ALG@printindent@tempcnta
\def\ALG@printindent{%
    \ifnum \theALG@nested>0
    \ifx\ALG@text\ALG@x@notext
    \else
    \unskip
    \ALG@printindent@tempcnta=1
    \loop
    \algrule[\csname ALG@ind@\the\ALG@printindent@tempcnta\endcsname]%
    \advance \ALG@printindent@tempcnta 1
    \ifnum \ALG@printindent@tempcnta<\numexpr\theALG@nested+1\relax
    \repeat
    \fi
    \fi
}%
\patchcmd{\ALG@doentity}{\noindent\hskip\ALG@tlm}{\ALG@printindent}{}{\errmessage{failed to patch}}
\AtBeginEnvironment{algorithmic}{\lineskip0pt}

\newcommand{\stdfont}[1]{ #1}

\begin{document}

\begin{frontmatter}

\title{Sauron U-Net: Simple automated redundancy elimination in medical image segmentation via filter pruning}

\author{Juan Miguel Valverde\textsuperscript{a}, Artem Shatillo\textsuperscript{b}, Jussi Tohka\textsuperscript{a}}
\address{\textsuperscript{a}A.I. Virtanen Institute for Molecular Sciences, University of Eastern Finland, 70150 Kuopio, Finland}
\address{\textsuperscript{b}Charles River Discovery Services, Kuopio 70210, Finland.}

\begin{abstract}
We introduce Sauron, a filter pruning method that eliminates redundant feature maps of convolutional neural networks (CNNs).
Sauron optimizes, jointly with the loss function, a regularization term that promotes feature maps clustering at each convolutional layer \hl{by reducing the distance between feature maps}. Sauron then eliminates the filters corresponding to the redundant feature maps by using automatically adjusted layer-specific thresholds.
Unlike most filter pruning methods, Sauron requires minimal changes to typical neural network optimization because it prunes and optimizes CNNs jointly, which, in turn, accelerates the optimization over time.
Moreover, unlike with other cluster-based approaches, the user does not need to specify the number of clusters in advance, a hyperparameter that is difficult to tune. We evaluated Sauron and five state-of-the-art filter pruning methods on four medical image segmentation tasks. This is an area where little attention has been paid to filter pruning, but \hl{where smaller CNN models are desirable for local deployment, mitigating privacy concerns associated with cloud-based solutions}.
\hl{Sauron was the only method that achieved a reduction in model size of over 90\% without deteriorating substantially the performance.}
\hl{Sauron also achieved, overall, the fastest models at inference time in machines with and without GPUs.}
Finally, we show through experiments that the feature maps of models pruned with Sauron are highly interpretable, which is essential for medical image segmentation.

\end{abstract}

\end{frontmatter}


\section{Introduction}

Pruning is a technique for eliminating unnecessary parameters from neural networks to obtain compact models and speed up their inference.
There are two main strategies for pruning convolutional neural networks (CNNs): weight pruning and filter pruning.
Weight pruning sets the weights of unimportant connections to zero \citep{lecun1990optimal,hassibi1993optimal,han2015deep,han2015learning,tung2018clip}, whereas filter pruning methods directly eliminate CNN filters.
Compared to weight pruning, the advantage of filter-pruned networks is that their training and inference requires no specialized hardware or software to handle sparse weights \citep{courbariaux2016binarized,rastegari2016xnor}.
There is a growing interest in applying pruning to medical imaging tasks, such as medical image segmentation.
This is because privacy concerns often prevent the transfer of medical images to cloud servers \citep{kaissis2020secure}, making small models that can be deployed locally essential.
Additionally, models with a few filters can be easier to interpret than large models, which is crucial not only in clinical but also in research applications.
There have been a number of recent studies that have applied filter pruning to medical image segmentation.
\cite{zhou2019evolutionary} used an evolutionary algorithm to prune CNNs for retinal vessel and neuronal membrane segmentation.
\cite{dinsdale2022stamp} proposed a novel pruning method called STAMP that was evaluated on MRI, CT, and ultrasound images.
\cite{jaiswal2023attend} used a technique called attention-based pruning to identify important channels for skin lesion segmentation.
\cite{lin2023lighter} proposed a pruning algorithm for vision transformers that was applied to segmentation of skin lesions from dermoscopy images and abdominal organs from CT.
Motivated by the above points and studies, we propose a filter pruning method called Sauron, and we demonstrate its application on U-Net-like networks \citep{ronneberger2015u}.

Most filter pruning frameworks consist of three distinct phases: Pre-training the model, pruning its filters, and fine-tuning or re-training to compensate for the loss of performance due to pruning \citep{liu2019metapruning,chang2020ucp}. 
Another class of methods combine pruning with training \citep{you2019gate,zhao2019variational,he2019filter,singh2019play} or fine-tuning \citep{luo2020autopruner,lin2018accelerating}, resulting in two-phase frameworks, and yet other methods repeat these phases multiple times \citep{you2019gate,luo2020autopruner,basha2021deep}.
Sauron, in contrast, applies filter pruning during optimization in a \textit{single phase}.
This strategy is advantageous compared to multiple-phases methods because it requires fewer hyperparameters and design decisions \citep{huang2018data}, such as the number of epochs for training and fine-tuning, pruning iterations, or whether to combine pruning with training or fine-tuning.
Additionally, Sauron does not insert new parameters into the optimized architecture to identify filter candidates for pruning, such as channel importance masks \citep{chang2020ucp,luo2020autopruner,hou2019weighted,lin2018accelerating,huang2018data}. This avoids potential optimization hindrance and requires less extra training time and GPU memory.

Sauron facilitates and promotes the formation of feature map clusters via a regularization term, and, unlike previous cluster-based approaches \citep{ding2019centripetal,he2019filter,basha2021deep}, Sauron does not enforce the number of the clusters.
Since the number of the clusters is expected to vary depending on the training data and across layers, the optimal number of feature maps per cluster is likely to differ.
Thus, determining the number of clusters is not trivial and may limit the accuracy and the pruning rate.

Our specific contributions are the following:

\begin{itemize}
\item We introduce Sauron, a single-phase filter pruning method that requires minimal changes to typical CNN optimization.
\item We propose a regularization term that promotes the formation of clusters of feature maps.
\item We show that Sauron provides both better test set performance, more compressed, and \hl{faster} models than competing methods in four medical image segmentation tasks.
\item We show that the feature maps generated by a model pruned with Sauron are highly interpretable.
\end{itemize}

\section{Related work} \label{sec:previouswork}

\subsection{Filter importance}
Most filter pruning approaches eliminate filters based on their importance.
Filter importance can be computed either after or during the optimization.
Then, the filters with the lowest importance score are eliminated.
Various measurements have been used to represent filter importance, including $L_p$ norms \citep{li2016pruning,xie2020localization,singh2019play,chen2022mtp,dinsdale2022stamp}, entropy \citep{luo2017entropy}, dot product \citep{he2021cap}, or post-pruning accuracy \citep{abbasi2017structural}.
Unimportant filters can also be identified via particle filtering \citep{anwar2017structured}, or genetic algorithms \citep{zhou2019evolutionary}.
Other methods add masks to each convolutional layer filter to learn channel importance \citep{chang2020ucp,luo2020autopruner,hou2019weighted,lin2018accelerating,huang2018data,ditschuneit2022auto}.
After training, these masks can be used to prune unimportant filters. 
However, channel importance masks modify the network architecture, which can increase GPU memory usage during training, prolong optimization time, and potentially impair the optimization.
Alternatively, other methods consider the scaling factor of batch normalization layers as channel importance \citep{ye2018rethinking,zhao2019variational}, but, in medical image segmentation, batch normalization is occasionally replaced by other normalization layers due to the small mini-batch size \citep{isensee2021nnu}.

\subsection{Difference minimization}
Difference minimization methods remove filters while trying to preserve characteristics of the original unpruned models.
Examples of these characteristics include classification accuracy \citep{liu2019metapruning}, Taylor-expansion-approximated loss \citep{you2019gate}, and the feature maps \citep{yu2018nisp,wang2018exploring,xie2020localization,luo2018thinet}.
A disadvantage of these methods is that they require a large GPU memory to avoid loading and unloading the unpruned models in memory constantly, which would slow down the training.
Furthermore, since finding the appropriate filters for their elimination is NP-hard, certain methods resorted to selecting filters based on their importance \citep{yu2018nisp,xie2020localization,you2019gate}, or via genetic \citep{liu2019metapruning} or greedy \citep{luo2018thinet} algorithms.

\subsection{Redundancy elimination}
Redundancy elimination approaches, including Sauron, eliminate redundant filters or feature maps. They identify redundant filters by computing a similarity metric among all \citep{wang2019cop,suau2020filter} or within clusters of filters/feature maps \citep{he2019filter,ding2019centripetal,basha2021deep}.
Previously, cluster-based approaches have considered redundant those within-cluster filters near the Euclidean center \citep{ding2019centripetal} and median \citep{he2019filter}, or filters with similar $L_1$ norm over several training epochs \citep{basha2021deep}.
A disadvantage of these approaches is the extra ``number of clusters" hyperparameter, which is data dependent and the same hyperparameter value might not be optimal across all layers.
Other methods used the Pearson's correlation between the weights \citep{wang2019cop} or between the feature maps \citep{suau2020filter} within the same layer, and feature maps' rank \citep{lin2020hrank} to indicate redundancy. However, their computations are more expensive than utilizing distances as in cluster-based methods.
More recently, hierarchical and spectral clustering \citep{tian2021energy,wang2022filter} has also been explored for filter pruning.

Concurrent with our work, \cite{dinsdale2022stamp} proposed a filter pruning method, called STAMP, for medical image segmentation.
Our method Sauron differs in several ways from STAMP.
First, Sauron, as detailed in \Cref{sec:sauron}, is a redundancy elimination method that increases the redundancy of feature maps through regularization, whereas STAMP is an importance-based method that computes the importance of feature maps using the $L_2$ norm.
Second, unlike STAMP, Sauron does not rely on a post-pruning step to recover the model from possible performance degradation, similar to the fine-tuning step in multi-stage filter pruning methods.
Finally, we evaluated Sauron on nnUNet, an advanced architecture trained with extensive data augmentation that has achieved the best segmentation performance on multiple datasets \citep{isensee2021nnu}.
In contrast, STAMP was applied to UNet \citep{ronneberger2015u}, a much older architecture.
We compare STAMP with Sauron in \Cref{sec:benchmark} since STAMP was originally evaluated on MRI and CT image segmentation as in our experiments.

\section{Sauron} \label{sec:sauron}

In this section, we present our approach to filter pruning, which we call \textbf{S}imple \textbf{AU}tomated \textbf{R}edundancy eliminati\textbf{ON} (Sauron).
Sauron optimizes, jointly with the loss function, a regularization term that promotes the clustering of feature maps at each convolutional layer, accentuating CNNs' redundancy.
It then eliminates the filters corresponding to the redundant feature maps by using automatically adjusted layer-specific thresholds. 
Sauron requires minimal changes to the typical neural network optimization since it prunes and optimizes CNNs jointly, i.e., training involves the usual forward-backward passes and a pruning step after each epoch.
Moreover, Sauron does not integrate optimizable parameters, such as channel importance masks \citep{chang2020ucp,luo2020autopruner,hou2019weighted,lin2018accelerating,huang2018data,ditschuneit2022auto}, into the CNN architecture.
This avoids complicating the optimization task and increasing the training time and the required GPU memory.
\cref{alg:sauron} summarizes our method.

\begin{algorithm}[t]
\caption{Sauron}
\label{alg:sauron}
\begin{algorithmic}[1]
\INPUT training data: $\mathcal{D}$.
\State  \textbf{Given}: $\lambda$, maximum threshold $\tau_{max}$, $epochs$, percentage of pruned filters $\mu$, patience $\rho$, number of steps $\kappa$.
\State  \textbf{Initialize}: model's weights $\mathbf{W} \leftarrow \{\mathbf{W} ^{l}, 1\leq l \leq L\}$, layer-specific thresholds $\boldsymbol{\tau} \leftarrow \{\tau_{l} = 0, 1\leq l \leq L\}$
\For{$e=1$; $e \leq epochs$}
	\For{$b=1$; $b \leq N $} \textit{\# Mini batches}
        \State \textit{\# Forward pass}
	    \State Compute predictions $\boldsymbol{\hat{y}}$, and loss $\mathcal{L}$
        \State Compute $\delta_{opt}$ (\cref{eq:deltaopt}), and $\boldsymbol{\delta}_{prune}$ (\cref{eq:grouprand})
        \State \textit{\# Backward pass}
	    \State  Update $\boldsymbol{\theta}$
	\EndFor

    \State \textit{\# Pruning step}
	\For{$l=1$; $l \leq L$}
	    \State \textit{\#\# Procedure 1: Increasing $\tau_l$ \#\#}
	    
	    \State C1: Training loss is converging
        \State C2: Validation loss is not improving
	    \State C3: Less than $\mu$\% of filters pruned in $(e-1)$
        \State C4: $\tau_l$ has not increased in last $\rho$ epochs
	    
	    \If{(C1 $\land$ C2 $\land$ C3 $\land$ C4) $\land$ ($\tau_l < \tau_{max}$) }
	        \State $\tau_l \leftarrow \tau_l + \tau_{max}/\kappa$
	    \EndIf
	    
	    \State \textit{\#\# Procedure 2: Pruning \#\#}
	    \State $\mathbf{W}^l \leftarrow \{\mathbf{W}^l : \boldsymbol{d}^l > \tau_l \}$
    \EndFor  
    
\EndFor
\OUTPUT Pruned CNN.
\end{algorithmic} 
\end{algorithm}

\subsection{Preliminaries}
Let $\mathcal{D} = \left\{\mathbf{x}_{i}, \boldsymbol{y}_{i}\right\}_{i=1}^{N}$ represent the training set, where $\mathbf{x}_{i} \in \mathbb{R}^{d}$ denotes image $i$, $\mathbf{y}_{i}$ its corresponding segmentation, and $N$ is the number of images.
$d$ is the image size that, in 2D images, corresponds to \textit{height} $\times$ \textit{width}, and, in 3D images, it corresponds to \textit{height} $\times$ \textit{width} $\times$ \textit{depth}.
Let $\mathbf{W}^l \in \mathbb{R}^{s_{l+1} \times s_l \times k^p}$ be the weights, composed by $s_{l+1} s_l$ filters of size $k^p$ at layer $l$, where $s_{l+1}$ denotes the number of output channels, $s_l$ the number of input channels, $k$ is the kernel size, and $p \in \{2,3\}$ is the number of dimensions.
Given feature maps $\mathbf{O}^l \in \mathbb{R}^{s_l \times d}$ at layer $l$ the feature maps $\mathbf{O}^{l+1} \in \mathbb{R}^{s_{l+1} \times d}$ at layer $l + 1$ are computed as
\begin{align} \label{eq:outputfilter}
    \mathbf{O}^{l+1} = \sigma(Norm(\mathbf{W}^l * \mathbf{O}^l)),
\end{align}
where * is the convolution operation, $Norm$ is a normalization layer, and $\sigma$ is an activation function.
For simplicity, we omit the bias term in \eqref{eq:outputfilter}, and we include all CNN's parameters in $\boldsymbol{\theta} = \{\boldsymbol{W}^1, \ldots, \boldsymbol{W}^L\}$, where $L$ is the number of layers.
We denote the predicted segmentation of the image $\mathbf{x}_{i}$ by $\boldsymbol{\hat{y}}_i$.

\subsection{Forward and backward pass: $\delta_{opt}$ regularization} \label{sec:forwardbackwardpass}
Sauron minimizes a loss $\mathcal{L}$ consisting of Cross Entropy $\mathcal{L}_{CE}$, Dice loss $\mathcal{L}_{Dice}$ \citep{milletari2016v}, and a novel channel distance regularization term $\delta_{opt}$: $\mathcal{L} =  \mathcal{L}_{CE} + \mathcal{L}_{Dice} + \lambda \delta_{opt}$, where
\begin{align} \label{eq:deltaopt}
    \delta_{opt} = \frac{1}{L} \sum_{l=1}^L \frac{1}{s_{l+1}} \sum_{r=1,r\neq t}^{s_{l+1}} || \phi(\boldsymbol{O}_t^l; \omega) - \phi(\boldsymbol{O}_{r}^l; \omega)||_2,
\end{align}
$\lambda$ is a hyperparameter that balances the contribution of $\delta_{opt}$, $\phi$ denotes average pooling with window size and strides $\omega$, and $t \in \{1,\ldots, s_{l+1}\}$ is an index to a randomly-chosen feature map.
\hl{A large $\omega$ will decrease the size of the feature maps and, consequently, speed up the computation of $\delta_{opt}$.}
$t$ may differ across layers but it is fixed during the optimization; since CNN weights are randomly initialized, the choice of $t$ is inconsequential and we set it to 1.
Before computing $\delta_{opt}$, feature maps $\boldsymbol{O}_t^l$ and $\boldsymbol{O}_{-t}^l$ (all channels except $t$) are normalized to the range $[0, 1]$ via min-max normalization, as we experimentally found this normalization strategy to be the best (see Section 1 of the Supplementary material).

Minimizing the proposed regularization term $\delta_{opt}$ further promotes the formation of feature maps clusters that already results from standard CNN optimization \citep{he2019filter,wang2019cop}.
In other words, the optimization of the loss $\mathcal{L}$ consists of 1) minimizing $\mathcal{L}_{CE} + \mathcal{L}_{Dice}$, that aims to solve the segmentation task, producing feature maps clusters, and 2) minimizing $\delta_{opt}$, that tends to align all feature maps $\boldsymbol{O}_{-t}^l$ to $\boldsymbol{O}_t^l$.
Similarly to typical $L_1$ and $L_2$ regularization, $\lambda$ limits the contribution of our regularization term $\delta_{opt}$, avoiding the complete alignment of all feature maps to $\boldsymbol{O}_t^l$ that is not desired.
During $\delta_{opt}$ minimization, the distance between $\boldsymbol{O}_t^l$ and those feature maps nearby, i.e., from the same cluster, is reduced.
At the same time, those feature maps in other clusters far from $\boldsymbol{O}_t^l$ become more similar to their neighbor feature maps, as it holds that $|| \phi(\boldsymbol{O}_{i}^l; \omega) - \phi(\boldsymbol{O}_{j}^l; \omega)||_2 \leq || \phi(\boldsymbol{O}_t^l; \omega) - \phi(\boldsymbol{O}_{i}^l; \omega)||_2 + || \phi(\boldsymbol{O}_t^l; \omega) - \phi(\boldsymbol{O}_{j}^l; \omega)||_2$ for $i \neq j$, i.e., the right hand side---minimized via $\delta_{opt}$ regularization---is an upper bound of the left hand side.
We demonstrate this clustering effect in \Cref{sec:fewclusters}.
Furthermore, we focus on pruning redundant feature maps rather than weights since different non-redundant weights can lead to similar feature maps.
In other words, eliminating redundant weights guarantees no reduction in feature maps redundancy.

\subsection{Pruning step} \label{sec:pruningstep}
Sauron computes the distances between a randomly-chosen feature map $\pi \in \{1,\ldots, s_{l+1}\}$ and all the others: $\boldsymbol{\delta}_{prune} = \{d^l_{r}/\max_rd^l_{r}:l = 1, \ldots, L, r = 1,\ldots, \pi - 1, \pi +1, \ldots,s_{l+1}\}$, where
\begin{align} \label{eq:grouprand}
    d^l_{r} = || \phi(\boldsymbol{O}_\pi^l; \omega) - \phi(\boldsymbol{O}_{r}^l; \omega)||_2.
\end{align}
Importantly, $\pi$ is different in every layer and epoch, enabling Sauron to prune different feature map clusters.
Moreover, since finding an appropriate pruning threshold requires the distances to lie within a known range, Sauron normalizes $d^l_{r}$ such that their maximum is $1$, i.e., $d^l_{r} \leftarrow d^l_{r} / \max_r(d^l_r)$.

Sauron employs layer-specific thresholds $\boldsymbol{\tau} = [\tau_1, \ldots, \tau_L]$, where all $\tau_l$ are initialized to zero and increase independently (usually at a different pace) until reaching $\tau_{max}$.
This versatility is important as the ideal pruning rate differs across layers due to their different purpose (i.e., extraction of low- and high-level features) and their varied number of filters.
Additionally, this setup permits utilizing high thresholds without removing too many filters at the beginning of the optimization, as feature maps may initially lie close to each other due to the random initialization.
In consequence, pruning is embedded into the training and remains \textit{always active}, portraying Sauron as a single-phase filter pruning method.

\paragraph{Procedure 1: Increasing $\tau_l$} Pruning with adaptively increasing layer-specific thresholds raises two important questions: how and when to increase the thresholds?
Sauron increases the thresholds linearly in $\kappa$ steps until reaching $\tau_{max}$.
\hl{A small $\kappa$ leads to faster and more aggressive pruning, while a large $\tau_{max}$ leads to larger pruning rates.}
Then, thresholds are updated once the model has stopped improving (C1 and C2 in \cref{alg:sauron}) and it has pruned only a few filters (C3).
An additional ``patience" hyperparameter ($\rho$) ensures that the thresholds are not updated consecutively (C4).
Conditions C1$, \dots, $C4 are easy to implement and interpret, and they rely on heuristics commonly employed for detecting convergence.

\paragraph{Procedure 2: Pruning} 
Sauron considers nearby feature maps to be redundant since they likely belong to the same cluster.
In consequence, Sauron removes all input filters $\mathbf{W}^l_{\cdot,s_l}$ whose corresponding feature map distances $\boldsymbol{\delta}_{prune}$ are lower than threshold $\tau_l$.
In contrast to other filter pruning methods, Sauron needs to store no additional information, such as channel indices, and the pruned models become more efficient \textit{and} smaller.
Additionally, since pruning occurs during training, Sauron accelerates the optimization of CNNs \hl{(see Section 2 of the Supplementary material)}.
After training, pruned models can be easily loaded by specifying the new post-pruning number of input and output filters in the convolutional layers.

\subsection{Implementation} \label{sec:implementation}
Sauron decreased feature maps dimensionality via average pooling with window size and stride $\omega = 2$, and utilized $\lambda=0.5$ in the loss, maximum pruning threshold $\tau_{max} = 0.3$, pruning steps $\kappa = 15$, and patience $\rho = 5$ (C4 in \cref{alg:sauron}).
Additionally, we employed simple conditions to detect convergence for increasing the layer-specific thresholds $\boldsymbol{\tau}$.
Convergence in the training loss (C1) was detected once the most recent training loss lay between the maximum and minimum values obtained during the training.
We considered that the validation loss stopped improving (C2) once its most recent value increased with respect to all previous values.
Finally, the remaining condition (C3) held true if the layer-specific threshold pruned less than 2\% of the filters pruned in the previous epoch, i.e., $\mu = 2$.

Sauron's simple design permits its incorporation into existing CNN optimization frameworks easily.
As an example, in our implementation, convolutional blocks are wrapped into a class that computes $\delta_{opt}$ and $\boldsymbol{\delta}_{prune}$ effortlessly in the forward pass, and the pruning step is a callback function triggered after each epoch.
This implementation, together with the code for running our experiments and processing the datasets, was written in Pytorch \citep{paszke2019pytorch} and is publicly available at \url{https://github.com/jmlipman/SauronUNet}.
In our experiments, we utilized an Nvidia GeForce GTX 1080 Ti (11GB), and a server with eight Nvidia A100 (40GB).

\section{Results} \label{sec:experiments}
In this section, we compare Sauron with other state-of-the-art filter pruning methods (\cref{sec:benchmark}).
We empirically demonstrate that the proposed $\delta_{opt}$ regularization increases feature map clusterability (\cref{sec:fewclusters}), and we conduct an ablation study to show the impact of $\delta_{opt}$ regularization on pruning and performance (\cref{sec:ablation}). Finally, we visualize the feature maps of Sauron-pruned models (\cref{sec:interpretation}).

\subsection{Datasets}
We employed four 3D medical image segmentation datasets: Rats, ACDC, ATLAS, and KiTS.
\textit{Rats} comprised 160 3D T2w MRIs of rat brains with lesions \citep{valverde2020ratlesnetv2}, and the segmentation task was separating lesion from non-lesion voxels.
We divided Rats dataset into 0.8:0.2 train-test splits, and the training set was further divided into a 0.9:0.1 train-validation split, resulting in 115, 13, and 32 images for training, validation, and test, respectively.
\textit{ACDC} included the Automated Cardiac Diagnosis Challenge 2017 training set \citep{bernard2018deep}, comprised by 200 3D MRIs of 100 individuals.
The segmentation classes were background, right ventricle (RV), myocardium (M), and left ventricle (LV).
We divided ACDC dataset similarly to Rats dataset, resulting in 144, 16, and 40 images for training, validation, and test, respectively.
We only utilized ACDC's competition training set due to the limitation to only four submissions to the online platform of ACDC challenge.
\textit{ATLAS} dataset included the 655 3D T1w MRIs of ATLAS R2.0 challenge training set \citep{liew2022large} with manually annotated brain lesions.
We only considered the training set due to current limitations in the challenge submission platform, and, we divided ATLAS dataset similarly to Rats and ACDC datasets, resulting in 471, 53, 131 images for training, validation, and test.
Finally, \textit{KiTS} was composed by 210 3D CT images from Kidney Tumor Challenge 2019 training set, segmented into background, kidney and kidney tumor \citep{heller2019kits19}.
KiTS training set was divided into a 0.9:0.1 train-validation split, resulting in 183 and 21 images for training and validation.
We report the results on the KiTS's competition test set (90 3D images).
All 3D images were standardized to zero mean and unit variance.
The train-validation-test divisions and computation of the evaluation criteria was at the subject level, ensuring that the data from a single subject was completely in the train set or in the test set, never dividing subject's data between train and test sets.

\subsection{Benchmark on three segmentation tasks} \label{sec:benchmark}

\begin{table}
\caption{Performance on Rats, ACDC, ATLAS, and KiTS datasets. \textbf{Bold}: best performance among pruning methods.}
\label{table:performance}
\centering
\small{
\begin{tabular}{c|cl|cc}
\hline
& & Method & Dice & HD95 (mm) \\
\hline
\parbox[t]{2mm}{\multirow{6}{*}{\rotatebox[origin=c]{90}{\shortstack[c]{Rats}}}} &
\parbox[t]{2mm}{\multirow{6}{*}{\rotatebox[origin=c]{90}{\shortstack[c]{Lesion}}}}
& nnUNet & 0.94 \stdfont{$\pm$ 0.03} & 1.1 \stdfont{$\pm$ 0.3} \\
& & Sauron & \textbf{0.94 \stdfont{$\pm$ 0.03}} & 1.1 \stdfont{$\pm$ 0.3} \\
& & cSGD ($r=0.5$) & 0.86 \stdfont{$\pm$ 0.13} & 9.6 \stdfont{$\pm$ 16.8} \\
& & FPGM ($r=0.5$) & 0.93 \stdfont{$\pm$ 0.04} & \textbf{0.5 \stdfont{$\pm$ 0.5}} \\
& & Autopruner & 0.91 \stdfont{$\pm$ 0.04} & 0.8 \stdfont{$\pm$ 1.2} \\
& & STAMP & 0.92 \stdfont{$\pm$ 0.04} & 0.6 \stdfont{$\pm$ 0.4} \\
& & TVSPrune & 0.92 \stdfont{$\pm$ 0.04} & 0.7 \stdfont{$\pm$ 1.0} \\
\hline
\parbox[t]{2mm}{\multirow{18}{*}{\rotatebox[origin=c]{90}{\shortstack[c]{ACDC}}}} &
\parbox[t]{2mm}{\multirow{6}{*}{\rotatebox[origin=c]{90}{\shortstack[c]{LV}}}}
& nnUNet & 0.91 \stdfont{$\pm$ 0.05} & 4.4 \stdfont{$\pm$ 3.0} \\
& & Sauron & \textbf{0.90 \stdfont{$\pm$ 0.06}} & \textbf{4.7 \stdfont{$\pm$ 3.2}} \\
& & cSGD ($r=0.5$) & 0.10 \stdfont{$\pm$ 0.15} & 72.6 \stdfont{$\pm$ 74.1} \\
& & FPGM ($r=0.5$) & 0.57 \stdfont{$\pm$ 0.13} & 37.8 \stdfont{$\pm$ 7.3} \\
& & Autopruner & 0.88 \stdfont{$\pm$ 0.07} & 5.9 \stdfont{$\pm$ 4.6} \\
& & STAMP & 0.82 \stdfont{$\pm$ 0.12} & 8.4 \stdfont{$\pm$ 4.5} \\
& & TVSPrune & 0.75 \stdfont{$\pm$ 0.20} & 4.9 \stdfont{$\pm$ 7.4} \\
\cline{2-5}
& \parbox[t]{2mm}{\multirow{6}{*}{\rotatebox[origin=c]{90}{\shortstack[c]{M}}}}
& nnUNet & 0.90 \stdfont{$\pm$ 0.02} & 3.4 \stdfont{$\pm$ 5.8} \\
& & Sauron & \textbf{0.90 \stdfont{$\pm$ 0.02}} & 3.6 \stdfont{$\pm$ 8.0} \\
& & cSGD ($r=0.5$) & 0.54 \stdfont{$\pm$ 0.19} & 19.5 \stdfont{$\pm$ 35.6} \\
& & FPGM ($r=0.5$) & 0.89 \stdfont{$\pm$ 0.03} & \textbf{2.2 \stdfont{$\pm$ 1.6}} \\
& & Autopruner & 0.88 \stdfont{$\pm$ 0.03} & 2.5 \stdfont{$\pm$ 1.7} \\
& & STAMP & 0.88 \stdfont{$\pm$ 0.03} & 3.8 \stdfont{$\pm$ 4.2} \\
& & TVSPrune & 0.79 \stdfont{$\pm$ 0.07} & 2.9 \stdfont{$\pm$ 6.6} \\
\cline{2-5}
& \parbox[t]{2mm}{\multirow{6}{*}{\rotatebox[origin=c]{90}{\shortstack[c]{RV}}}}
& nnUNet & 0.95 \stdfont{$\pm$ 0.03} & 2.5 \stdfont{$\pm$ 1.8} \\
& & Sauron & \textbf{0.95 \stdfont{$\pm$ 0.03}} & 2.7 \stdfont{$\pm$ 2.0} \\
& & cSGD ($r=0.5$) & 0.64 \stdfont{$\pm$ 0.20} & 13.9 \stdfont{$\pm$ 8.2} \\
& & FPGM ($r=0.5$) & 0.00 \stdfont{$\pm$ 0.00} & 194.1 \stdfont{$\pm$ 23.5} \\
& & Autopruner & \textbf{0.95 \stdfont{$\pm$ 0.03}} & 3.1 \stdfont{$\pm$ 3.0} \\
& & STAMP & 0.94 \stdfont{$\pm$ 0.03} & 4.4 \stdfont{$\pm$ 6.7} \\
& & TVSPrune & 0.89 \stdfont{$\pm$ 0.07} & \textbf{1.4 \stdfont{$\pm$ 0.5}} \\
\hline
\parbox[t]{2mm}{\multirow{6}{*}{\rotatebox[origin=c]{90}{\shortstack[c]{ATLAS}}}} &
\parbox[t]{2mm}{\multirow{6}{*}{\rotatebox[origin=c]{90}{\shortstack[c]{Lesion}}}}
& nnUNet & 0.66 \stdfont{$\pm$ 0.25} & 17.7 \stdfont{$\pm$ 21.6} \\
& & Sauron & \textbf{0.66 \stdfont{$\pm$ 0.23}} & 17.8 \stdfont{$\pm$ 24.6} \\
& & cSGD ($r=0.5$) & 0.20 \stdfont{$\pm$ 0.22} & 86.0 \stdfont{$\pm$ 22.5} \\
& & FPGM ($r=0.5$) & 0.65 \stdfont{$\pm$ 0.24} & \textbf{16.3 \stdfont{$\pm$ 20.3}} \\
& & Autopruner & 0.54 \stdfont{$\pm$ 0.26} & 20.9 \stdfont{$\pm$ 21.0} \\
& & STAMP & 0.40 \stdfont{$\pm$ 0.25} & 29.5 \stdfont{$\pm$ 24.1} \\
& & TVSPrune & 0.01 \stdfont{$\pm$ 0.02} & 134.2 \stdfont{$\pm$ 16.7} \\
\hline
\end{tabular}
}
\vspace{0.3cm}

\centering
\small{
\begin{tabular}{c|cl|cc}
\hline
& & \multirow{2}{*}{Method} & Kidney & Tumor \\
& &  & Dice & Dice \\
\hline
\parbox[t]{2mm}{\multirow{6}{*}{\rotatebox[origin=c]{90}{\shortstack[c]{KiTS}}}} &
 & nnUNet & 0.9595 & 0.7657 \\
& & Sauron & \textbf{0.9564} & \textbf{0.7482} \\
& & cSGD ($r=0.5$) & 0.9047 & 0.5207 \\
& & FPGM ($r=0.5$) & 0.9509 & 0.6830 \\
& & Autopruner & 0.9167 & 0.5854  \\
& & STAMP & 0.8635 & 0.2183 \\
& & TVSPrune & 0.1148 & 0.0022 \\
\hline
\end{tabular}
}
\label{tab1}
\end{table}

\begin{table*}[t]
\caption{\hl{Decrease in FLOP, and decrease in GPU and CPU time at inference time with respect to the baseline nnUNet. Negative numbers indicate increase. \textbf{Bold}: highest decrease.}}
\label{table:compression}
\centering
\scriptsize{
\begin{tabular}{l|ccc|ccc|ccc|ccc}
\hline
\multirow{2}{*}{Method} & \multicolumn{3}{c}{Rats} & \multicolumn{3}{c}{ACDC} & \multicolumn{3}{c}{ATLAS} & \multicolumn{3}{c}{KiTS} \\
\cline{2-13}
  & FLOP & GPU & CPU & FLOP & GPU & CPU & FLOP & GPU & CPU & FLOP & GPU & CPU \\
\hline 
Sauron & \textbf{96.45\%} & \textbf{5.1}\% & \textbf{64.6}\% & \textbf{92.41\%} & \textbf{20.8}\% & \textbf{69.5}\% & \textbf{97.39\%} & \textbf{39.6}\% & \textbf{49.2}\% & \textbf{93.02\%} & 6.6\% & \textbf{47.4}\% \\
cSGD ($r=0.5$) & 50.03\% & -5.1\% & 18.8\% & 49.80\% & 14.5\% & 29.5\% & 49.81\% & 27.4\% & 19.8\% & 49.81\% & 5.8\% & 19.6\% \\
FPGM ($r=0.5$) & 50.00\% & -6.2\% & -2.6\% & 50.00\% & 10.4\% & -12.8\% & 50.0\% & -3.4\% & -1.1\% & 49.98\% & -1.6\% & -0.4\% \\
Autopruner & 83.61\% & -338\% & -612\% & 88.52\% & -660\% & -447\% & 77.14\% & -3314\% & -92.2\% & 82.00\% & -1478\% & -90.8\% \\
STAMP & 88.72\% & -30.2\% & 31.3\% & 89.21\% & 3.0\% & 32.7\% & 89.63\% & 12.6\% & 9.6\% & 88.24\% & \textbf{69.6}\% & 14.3\% \\
TVSPrune & 3.07\% & -7.1\% & -0.2\% & 19.34\% & 14.6\% & 14.7\% & 9.54\% & 9.7\% & -0.5\% & 27.14\% & 4.9\% & -0.3\% \\
\hline
\end{tabular}
}
\end{table*}

We optimized and pruned nnUNet \citep{isensee2021nnu}, whose architecture slightly differed on the number of filters, encoder-decoder levels, and normalization layer based on the dataset.
For Rats and ACDC, we utilized 2D nnUNet, and for ATLAS and KiTS, we utilized 3D nnUNet; further details related to nnUNet's architecture and its optimization can be found in Section 3 of the Supplementary material.
We compared Sauron's performance with cSGD \citep{ding2019centripetal}, FPGM \citep{he2019filter}, Autopruner \citep{luo2020autopruner}, STAMP \citep{dinsdale2022stamp}, and \hl{TVSPrune} \citep{murti2023tvsprune} using a pruning rate similar to the one achieved by Sauron.
We chose cSGD and FPGM since, similarly to Sauron, they are redundancy-elimination filter pruning methods.
Autopruner and STAMP, in contrast, are filter-importance-based methods; Autopruner learns filter importance during the training, and STAMP computes filter importance via $L_2$ norm.
We chose to compare STAMP against Sauron because it was also evaluated on medical image segmentation tasks.
\hl{We also evaluated TVSPrune, which can prune without access to the training data.}
Since cSGD and FPGM severely underperformed when adjusted to achieve a pruning rate similar to Sauron, we re-run them with their pruning rate set to only 50\% ($r=0.5$).
\hl{For TVSPrune, we aimed to avoid severe underperformance by limiting the number of pruning iterations.}
We computed the Dice coefficient \citep{dice1945measures} and 95\% Hausdorff distance (HD95) \citep{rote1991computing} on Rats, ACDC and ATLAS test sets (see \Cref{table:performance}).
In KiTS dataset, only the average Dice coefficient was provided by the online platform that evaluated the test set.
We computed the relative decrease in the number of floating point operations (FLOP) in all convolutions: $FLOP = HW(C_{in}C_{out})K^2$, where $H, W$ is the height and width of the feature maps, $C_{in}, C_{out}$ is the number of input and output channels, and $K$ is the kernel size.
For the 3D CNNs (ATLAS and KiTS datasets), we added extra multipliers $D$ (depth) and $K$ to compute the FLOP.
\hl{Additionally, we computed the average decrease (or increase) in inference time of the pruned networks with respect to the baseline nnUNet using only CPU and GPU. For this, we averaged the inference time of each method by dividing their total inference time by the number of images in each dataset. Then, we divided the average inference time of the pruned networks by the average inference time of the baseline nnUNet.}

Sauron obtained the highest Dice coefficients and competitive HD95s across all datasets and segmentation classes (\Cref{table:performance}).
Sauron also achieved the highest reduction in FLOP, although, every method, including Sauron, can further reduce the FLOP at the risk of worsening the performance (\Cref{table:compression}).
\hl{Similarly, Sauron also achieved the highest decrease in GPU and CPU time across all datasets, with the exception of KiTS, where STAMP produced a model that was faster if run on a GPU (but not on CPU).}
cSGD and FPGM could not yield models with high pruning rates possibly because \hl{they were designed to reduce only $s_{l+1}$ and not $s_l$ from $\mathbf{W}^l \in \mathbb{R}^{s_{l+1} \times s_l \times k \times k}$}.
Thus, very high pruning rates cause a great imbalance between the number of input and output filters in every layer that may hinder the training.
Note also that cSGD and FPGM were not tested with pruning rates higher than 60\% \citep{ding2019centripetal,he2019filter}.
In contrast, Sauron, Autopruner and STAMP, that achieved working models with higher pruning rate, reduced both input filters $s_{l}$ and output filters $s_{l+1}$.
\hl{TVSPrune did not achieve models with performance or compression rates similar to other methods likely because, unlike in }\cite{murti2023tvsprune}\hl{, we reserved the test set exclusively for evaluating the models and not for determining which filters to prune.}
We included an example segmentation from each dataset in Section 4 of the Supplementary material, and \hl{Sauron's training and validation loss in Section 5 of the Supplementary material}.

\begin{figure*}
\centering
    \includegraphics[width=\textwidth]{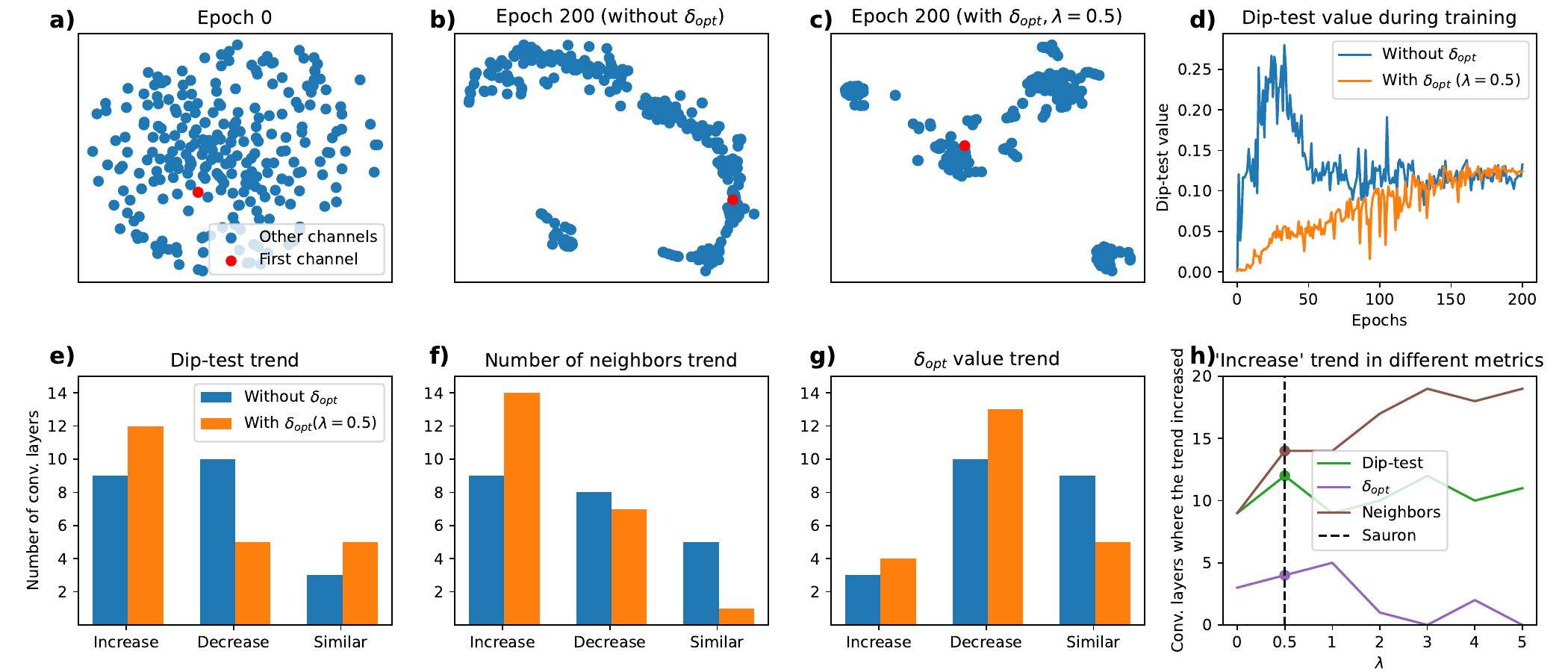} 
   \caption{a-c) tSNE plot of the feature maps of the first block of the decoder at initialization (epoch 0), and after optimizing with and without $\delta_{opt}$. d) Corresponding dip-test values during the optimization. e-g) Summary of the trends across the three clusterability measures in all convolutional layers. h) Number of layers with an increasing trend in the three clusterability measures with higher values of $\lambda$ (dashed line: Sauron's default configuration).} \label{fig:clusterability}
\end{figure*}

\subsection{Minimizing $\delta_{opt}$ promotes the formation of feature maps clusters} \label{sec:fewclusters}

We investigated feature map clustering tendency during nnUNet's optimization.
For this, we deactivated Sauron's pruning step and optimized $\mathcal{L}$ on Rats dataset with and without $\delta_{opt}$ while storing the feature maps at each epoch (including at epoch 0, before the optimization) of every convolutional layer.
Since quantifying clusterability is a hard task, we utilized three different measures:
1) We employed \textbf{dip-test} \citep{kalogeratos2012dip}, as Adolfsson \textit{et al.} \citep{adolfsson2019cluster} demonstrated its robustness compared to other methods for quantifying clusterability.
High dip-test values signal higher clusterability.
2) We computed the average \textbf{number of neighbors} of each feature map layer-wise.
Specifically, we counted the feature maps within $r$, where $r$ corresponded to the 20\% of the distance between the first channel and the farthest channel.
Distance $r$ is computed every time since the initial distance between feature maps is typically reduced while training.
An increase in the average number of neighbors indicates that feature maps have become more clustered.
3) We calculated the \textbf{average distance} to the first feature map channel (i.e., $\delta_{opt}$) for each layer, which illustrates the total reduction of those distances achieved during and after the optimization.

In agreement with the literature \citep{he2019filter,wang2019cop}, \cref{fig:clusterability} shows that optimizing nnUNet (without $\delta_{opt}$ regularization) yields clusters of feature maps.
Feature maps in the first block of the nnUNet decoder show no apparent structure suitable for clustering at initialization (\cref{fig:clusterability}, a), and, at the end of the optimization, feature maps appear more clustered (\cref{fig:clusterability}, b).
\cref{fig:clusterability} (d, blue line) also illustrates this phenomenon: dip-test value is low in the beginning and higher at the end of the training.
However, this increasing trend did not occur in all layers.
To illustrate this, we compared, for each layer, the average dip-test value, number of neighbors, and distance $\delta_{opt}$ in the first ($p_1$) and last third ($p_2$) of the training.
Then, we considered the trend similar if $|p_2-p_1| < 0.001$ for the dip-test values or $|p_2-p_1| < 0.05p_1$ for the number of neighbors and distance $\delta_{opt}$.
\cref{fig:clusterability} (e) shows that the number of layers in which the dip-test value increased and decreased were similar when not minimizing the $\delta_{opt}$ regularization term.
In contrast, the number of layers with an increasing trend was proportionally larger with $\delta_{opt}$ regularization.
\cref{fig:clusterability} (f) shows a similar outcome regarding the average number of neighbors, i.e., $\delta_{opt}$ regularization led to proportionally more neighbors near each feature map.
In the same line, the average distance between the first feature map and the rest decreased more with $\delta_{opt}$ regularization (\cref{fig:clusterability}, (f)).
Additionally, \cref{fig:clusterability} (c) also illustrates that incorporating the $\delta_{opt}$ regularization term enhances the clustering of feature maps, as there are more clusters and the feature maps are more clustered than when not minimizing $\delta_{opt}$ (\cref{fig:clusterability} (b)).

We observed higher clusterability in the convolutional layers with more feature maps (see Section 6 of the Supplementary material), likely because such convolutional layers contribute more to the value of $\delta_{opt}$.
On the other hand, convolutional layers with fewer feature maps have larger feature vectors (e.g., \textit{enc\_block\_1} feature vectors are $(256 \times 256) \times 32$ in Rats dataset) whose distances tend to be larger due to the curse of dimensionality.
Sauron accounts, to some extent, for these differences in the convolutional layers with the adaptively-increasing layer-specific thresholds $\boldsymbol{\tau}$.
Another possible way to tackle these differences is by using different layer-specific $\lambda$'s to increase the contribution of the distances of certain layers.
We investigated the impact on feature map clusterability with higher $\lambda$ values and, as illustrated in \cref{fig:clusterability} (h), a higher $\lambda$ tended to increase the average number of neighbors, decrease $\delta_{opt}$, and somewhat increase the dip-test values, which, overall, signals higher clusterability.

\subsection{Ablation study} \label{sec:ablation}

\begin{table}[t]
\caption{Performance and decrease in FLOP. \textbf{Bold:} Sauron's default $\lambda$.}
  \label{table:ablation}
  \centering
\footnotesize{
\begin{tabular}{c|cl|ccc}
\hline
& & Sauron's $\lambda$ & Dice & HD95 (mm) & FLOP $\downarrow$ \\
\hline
\parbox[t]{2mm}{\multirow{4}{*}{\rotatebox[origin=c]{90}{\shortstack[c]{Rats}}}} &
\parbox[t]{2mm}{\multirow{4}{*}{\rotatebox[origin=c]{90}{\shortstack[c]{Lesion}}}}
& $\lambda=0$ &  0.93 \stdfont{$\pm$ 0.03} & 1.2 \stdfont{$\pm$ 0.5} & 96.62\% \\
& & \boldsymbol{$\lambda=0.5$} &  0.94 \stdfont{$\pm$ 0.03} & 1.1 \stdfont{$\pm$ 0.3} & 96.45\% \\
& & $\lambda=1$ & 0.94 \stdfont{$\pm$ 0.03} & 0.4 \stdfont{$\pm$ 0.2} & 97.44\% \\
& & $\lambda=2$ & 0.94 \stdfont{$\pm$ 0.03} & 0.4 \stdfont{$\pm$ 0.2} & 97.43\% \\
\hline
\parbox[t]{2mm}{\multirow{12}{*}{\rotatebox[origin=c]{90}{\shortstack[c]{ACDC}}}} &
\parbox[t]{2mm}{\multirow{4}{*}{\rotatebox[origin=c]{90}{\shortstack[c]{LV}}}}
& $\lambda=0$ & 0.89 \stdfont{$\pm$ 0.08} & 5.3 \stdfont{$\pm$ 4.4} & 89.04\% \\
& & \boldsymbol{$\lambda=0.5$} & 0.90 \stdfont{$\pm$ 0.06} & 4.7 \stdfont{$\pm$ 3.2} & 92.41\% \\
& & $\lambda=1$ & 0.90 \stdfont{$\pm$ 0.06} & 5.1 \stdfont{$\pm$ 4.2} & 94.86\% \\
& & $\lambda=2$ & 0.90 \stdfont{$\pm$ 0.06} & 5.3 \stdfont{$\pm$ 3.5} & 95.20\% \\
\cline{2-6}
& \parbox[t]{2mm}{\multirow{4}{*}{\rotatebox[origin=c]{90}{\shortstack[c]{M}}}}
& $\lambda=0$ & 0.90 \stdfont{$\pm$ 0.02} & 2.4 \stdfont{$\pm$ 1.7} & 89.04\% \\
& & \boldsymbol{$\lambda=0.5$} & 0.90 \stdfont{$\pm$ 0.02} & 3.6 \stdfont{$\pm$ 8.0} & 92.41\% \\
& & $\lambda=1$ & 0.90 \stdfont{$\pm$ 0.02} & 2.3 \stdfont{$\pm$ 1.7} & 94.86\% \\
& & $\lambda=2$ & 0.90 \stdfont{$\pm$ 0.02} & 2.4 \stdfont{$\pm$ 1.7} & 95.20\% \\
\cline{2-6}
& \parbox[t]{2mm}{\multirow{4}{*}{\rotatebox[origin=c]{90}{\shortstack[c]{RV}}}}
& $\lambda=0$ & 0.95 \stdfont{$\pm$ 0.03} & 3.1 \stdfont{$\pm$ 3.0} & 89.04\% \\
& & \boldsymbol{$\lambda=0.5$} & 0.95 \stdfont{$\pm$ 0.03} & 2.7 \stdfont{$\pm$ 2.0} & 92.41\% \\
& & $\lambda=1$ & 0.95 \stdfont{$\pm$ 0.03} & 2.8 \stdfont{$\pm$ 2.0} & 94.86\% \\
& & $\lambda=2$ & 0.95 \stdfont{$\pm$ 0.03} & 2.7 \stdfont{$\pm$ 2.1} & 95.20\% \\
\hline
\parbox[t]{2mm}{\multirow{4}{*}{\rotatebox[origin=c]{90}{\shortstack[c]{ATLAS}}}} &
\parbox[t]{2mm}{\multirow{4}{*}{\rotatebox[origin=c]{90}{\shortstack[c]{Lesion}}}}
& $\lambda=0$ & 0.66 \stdfont{$\pm$ 0.22} & 17.6 \stdfont{$\pm$ 22.9} & 96.82\% \\
& & \boldsymbol{$\lambda=0.5$} & 0.66 \stdfont{$\pm$ 0.23} & 17.8 \stdfont{$\pm$ 24.6} & 97.39\% \\
& & $\lambda=1$ & 0.65 \stdfont{$\pm$ 0.24} & 17.0 \stdfont{$\pm$ 20.9} & 97.57\% \\
& & $\lambda=2$ & 0.64 \stdfont{$\pm$ 0.24} & 18.7 \stdfont{$\pm$ 23.9} & 99.27\% \\
\hline
\end{tabular}
}
\vspace{0.3cm}

\begin{tabular}{c|cl|ccc}
\hline
& & \multirow{2}{*}{Sauron's $\lambda$} & Kidney & Tumor & \multirow{2}{*}{FLOP $\downarrow$} \\
& &  & Dice & Dice & \\
\hline
\parbox[t]{2mm}{\multirow{4}{*}{\rotatebox[origin=c]{90}{\shortstack[c]{KiTS}}}} &
& $\lambda=0$ & 0.9556 & 0.7352 & 85.82\% \\
& & \boldsymbol{$\lambda=0.5$} & 0.9564 & 0.7482 & 93.02\% \\
& & $\lambda=1$ & 0.9572 & 0.7473 & 95.19\% \\
& & $\lambda=2$ & 0.9496 & 0.7427 & 97.22\% \\
\hline
\end{tabular}
\end{table}

We conducted ablation experiments with different $\lambda$ values to understand the influence of the proposed $\delta_{opt}$ regularization on performance and pruning rate.
\Cref{table:ablation} shows the  Dice coefficient, HD95, and decrease in FLOP with respect to the baseline nnUNet.
Sauron with $\delta_{opt}$ regularization (\Cref{table:ablation}, $\lambda>0$) produced more compressed models than without $\delta_{opt}$ while achieving similar Dice coefficients and HD95.
These results show that typical CNN optimization ($\lambda=0$) yields redundant feature maps that can be pruned with Sauron, and that pruning without $\delta_{opt}$ regularization can affect performance (\Cref{table:ablation}, Rats and KiTS, $\lambda=0$), possibly due to the accidental elimination of non-redundant filters.
Furthermore, the increase in model compression by only increasing $\lambda$ demonstrates that $\lambda$ effectively controls how much feature maps can be clustered.

\subsection{Feature maps interpretation} \label{sec:interpretation}
\begin{figure*}
\centering
    \includegraphics[width=0.4\textwidth]{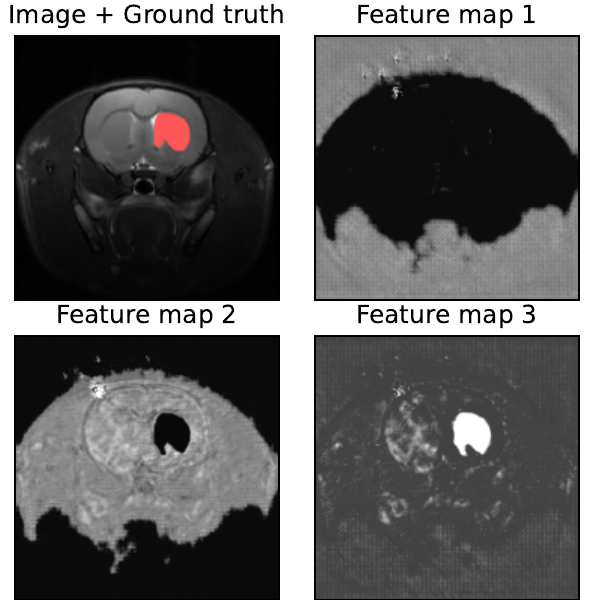}
    \includegraphics[width=0.4\textwidth]{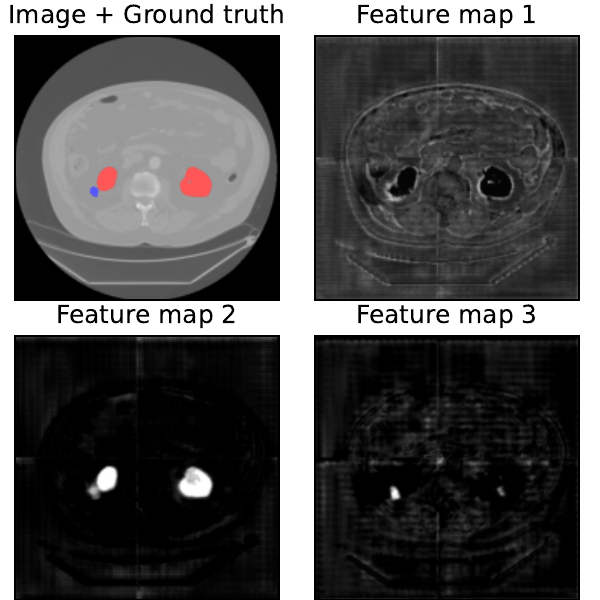}
    \includegraphics[width=0.6\textwidth]{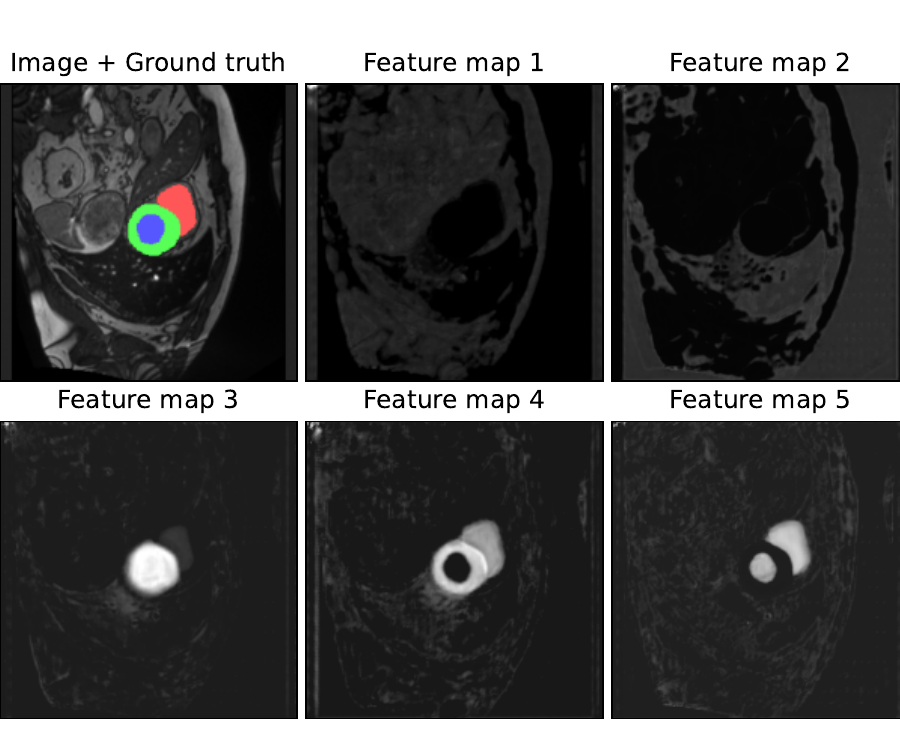}
    \includegraphics[width=0.6\textwidth]{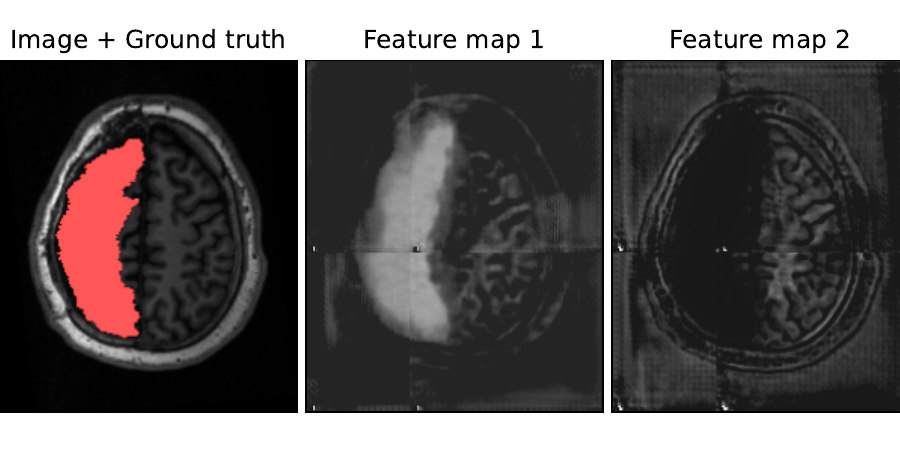}
   \caption{Image slice from Rats (top-left), KiTS (top-right), ACDC (middle), and ATLAS (bottom) datasets, its ground-truth segmentation, and all feature maps at the second-to-last convolutional block after pruning with Sauron.} \label{fig:featuremaps}
\end{figure*}

Sauron produces small and efficient models that can be easier to interpret.
This is due to $\delta_{opt}$ regularization that, as we showed in \Cref{sec:fewclusters,sec:ablation}, increases feature maps clusterability.
Each feature maps cluster can be thought of as a semantic operation and the cluster's feature maps as noisy outputs of such operation.
To test this view, we inspected the feature maps from the second-to-last convolutional block of a Sauron-pruned nnUNet (\Cref{fig:featuremaps}).
For comparison, the feature maps from the same convolutional layer of the baseline (unpruned) nnUNet are displayed in Section 7 of the Supplementary material.

The first feature map depicted in \cref{fig:featuremaps} (Rats dataset, top-left) captured the background and part of the rat head that does not contain brain tissue.
The second feature map contained the rest of the rat head without brain lesion, and the third feature map mostly extracted the brain lesion.
Although the third feature map seems to suffice for segmenting the brain lesion, the first feature map might have helped the model by discarding the region with no brain tissue at all.
In \cref{fig:featuremaps} (KiTS dataset, top-right), we can see that each feature map captured the background, kidney (red), and tumor (blue) with different intensities.
Similarly, the first and second feature maps in \cref{fig:featuremaps} (ACDC dataset, middle) detected the background, whereas feature maps 3, 4, and 5 extracted, with different intensities, the right cavity (red), myocardium (green), and left cavity (blue) of the heart.
The feature maps in \cref{fig:featuremaps} (ATLAS dataset, bottom) also show the location of the lesion.
The unpruned networks, on the other hand, not only yielded more feature maps but also there appear to be more clusters than in Sauron-pruned networks (see Section 7 of the Supplementary material).
The high-level interpretation of Sauron-derived feature maps facilitates understanding the role of the last convolutional block which, in the illustrated cases, could be replaced by simple binary operations.
This shows the interpretability potential of feature map redundancy elimination methods such as Sauron.

\section{Discussion} \label{sec:discussion}
Sauron achieved higher pruning rate, Dice coefficients and competitive HD95 compared to the evaluated methods \citep{ding2019centripetal,he2019filter,luo2020autopruner,dinsdale2022stamp}, even without $\delta_{opt}$ regularization (\Cref{table:performance,table:ablation}) \hl{and without optimizing Sauron's hyper-parameters ($\omega, \tau_{max}, \kappa, \rho$) on each dataset}, demonstrating that Sauron's pruning strategy provides state-of-the-art pruning rate and performance.
\hl{Overall, Sauron also achieved the largest decrease in FLOP and in CPU and GPU time at inference time. Our experiments comparing FLOP, CPU and GPU time showed that these three characteristics are not correlated, i.e., model compression and model speed at inference time may differ, likely because model speed depends on the hardware and software. Furthermore, certain pruning methods, such as Autopruner, produced models slower than the baseline because they incorporate additional layers to the models.}
Since Sauron is a single-phase \citep{zhao2019variational,dinsdale2022stamp} filter pruning method whose optimization resembles typical CNN training, it incorporates no design-related hyperparameters or decisions such as how to reconstruct feature maps \citep{wang2018exploring}, when and how many times distinct phases need to be executed \citep{you2019gate}, or whether to combine pruning with training \citep{you2019gate,zhao2019variational,he2019filter,singh2019play} or fine-tuning \citep{luo2020autopruner,lin2018accelerating}.
Additionally, single-phase methods such as Sauron can train, prune and fine-tune randomly-initialized models \textit{as well as} pretrained models.
In the medical domain, training from scratch is a common practice, and fine-tuning ImageNet-pretrained networks has been shown to not always be beneficial \citep{valverde2021transfer,raghu2019transfusion}.
Thus, the capability to prune randomly-initialized models is advantageous.
Furthermore, Sauron is not specific to any particular domain, task, or architecture.
This means that Sauron can be used in a variety of different applications, including natural image classification/segmentation and regression. Sauron can also be used with different neural network architectures, such as convolutional neural networks (CNNs), fully connected neural networks, and vision transformers \citep{dosovitskiy2020image}.
Sauron, however, cannot enforce specific compression rates because its pruning process is based on thresholding the distances between feature maps.
Additionally, since Sauron is data-driven, it requires access to the training data in order to prune the models.

We demonstrated that trained CNNs yield redundant feature maps (\cref{fig:clusterability}, b,d-g), which motivated this and previous works \citep{lin2020hrank,suau2020filter}. 
Additionally, using three different clusterability metrics and an ablation study, we demonstrated that $\delta_{opt}$ regularization increased the redundancy of feature maps  (see \Cref{sec:fewclusters,sec:ablation}).
Moreover, our ablation experiments showed that we could achieve higher compression rates by increasing $\lambda$ without sacrificing performance (\Cref{table:ablation}).
Our $\delta_{opt}$ regularization is in line with previous works that have used regularization methods to aid model compressibility \citep{basha2021deep,chen2022mtp,luo2020autopruner,huang2018data,suau2020filter,lin2020hrank}.
In particular, we showed that Sauron can exploit more clusterable feature maps, achieving higher compression (\Cref{table:compression}, ACDC, ATLAS, and KiTS datasets) and similar or better performance (\Cref{table:performance}, Rats (Lesion), ACDC (LV), KiTS (Kidney, Tumor) than without $\delta_{opt}$ regularization.

\section{Conclusion} \label{sec:conclusion}
We presented a single-phase filter pruning method named Sauron. We evaluated it on four medical image segmentation tasks, where Sauron yielded pruned models that were superior to the compared methods in terms of performance, pruning rate, and \hl{inference time}.
Our experiments demonstrated that $\delta_{opt}$ regularization increased the clusterability of feature maps, and that promoting such increase in feature maps clusterability is advantageous for filter pruning.
Finally, we showed that the few feature maps after pruning nnUNet with Sauron were highly interpretable.

\section*{Acknowledgments}
The work of J.M. Valverde was funded from the European Union’s Horizon 2020 Framework Programme (Marie Skłodowska Curie grant agreement \#740264 (GENOMMED)). This work was supported by the Research Council of Finland through grants 358944 (Flagship of Advanced Mathematics for Sensing Imaging and Modelling) and 346934. This work utilized the computational resources provided by UEF Bioinformatics Center, University of Eastern Finland, Finland.

\bibliography{mybibfile}

\newpage
\onecolumn

{\Large Supplementary Material}

\setcounter{section}{0}

\section{Normalization strategies} \label{appendix:normalization}

\begin{figure}[!htb]
\centering
    \includegraphics[width=0.8\textwidth]{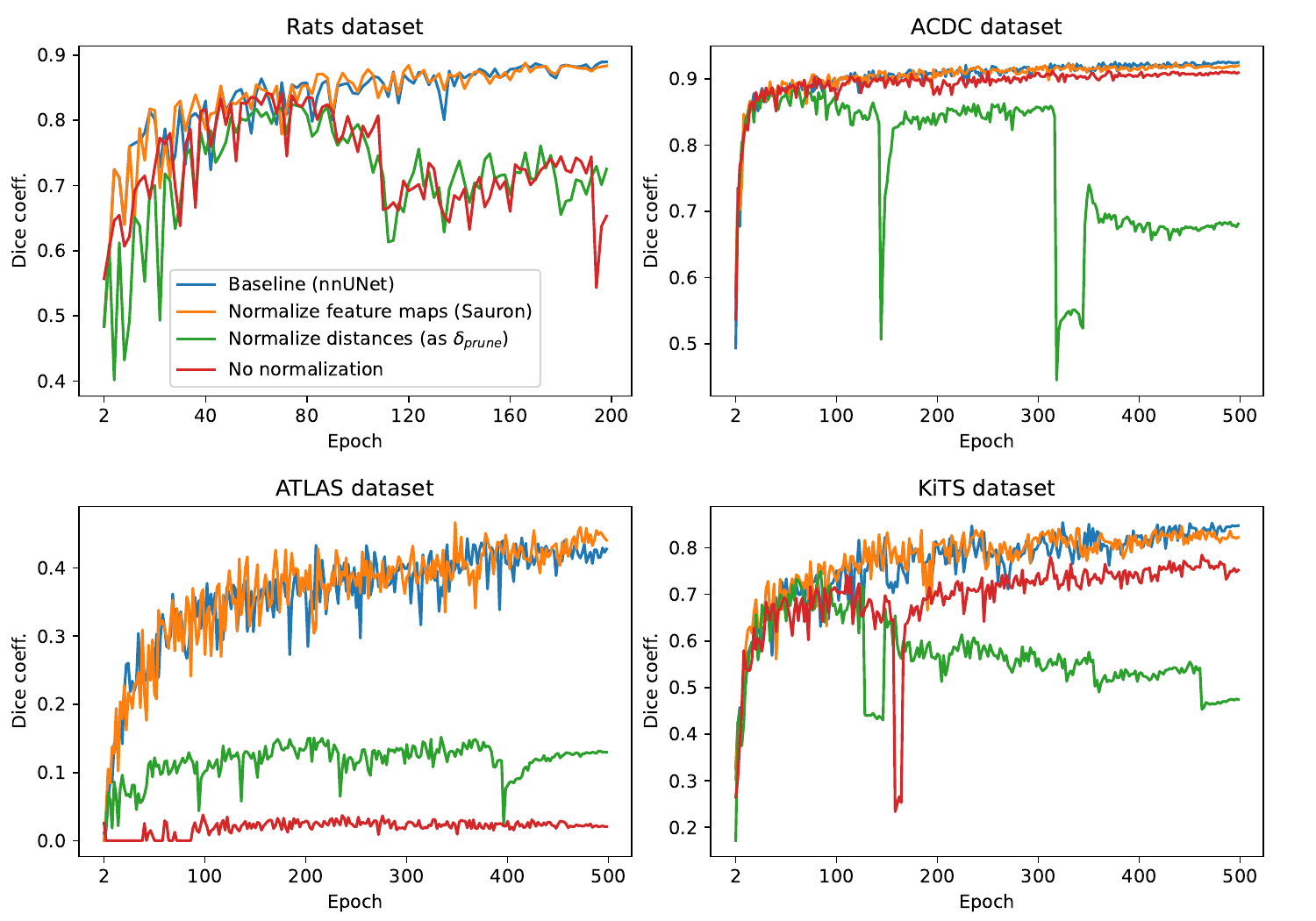} 
   \caption{Validation Dice coefficients of baseline nnUNet, Sauron, and two other approaches to normalize $\delta_{opt}$.} \label{fig:norm}
\end{figure}

We studied the impact on performance of different strategies to normalize $\delta_{opt}$.
For this, we compared Sauron's normalization strategy (orange, \cref{fig:norm}), the baseline nnUNet (blue), and when not normalizing $\delta_{opt}$ (red).
Additionally, we normalized $\delta_{opt}$ as in $\boldsymbol{\delta}_{prune}$ (green), i.e., instead of normalizing feature maps, the computed distances are divided by their maximum value, layer-wise (see Section 3.2).

\Cref{fig:norm} shows that Sauron's normalization of the feature maps provided the closest optimization stability and performance to the baseline nnUNet.

\clearpage
\newpage

\section{Sauron accelerates the optimization} \label{appendix:sauronaccelerates}
Since Sauron prunes filters during training, the optimization is faster than when not using Sauron.
Here, for each dataset, we computed the difference in time between consecutive epochs and divided such difference by the time taken during the first epoch. While nnUNet is expected to take the same time to compute each epoch, when pruning with Sauron, the epoch time decreases with respect to the first epoch.

\begin{figure}[!htb]
\centering
    \includegraphics[width=0.8\textwidth]{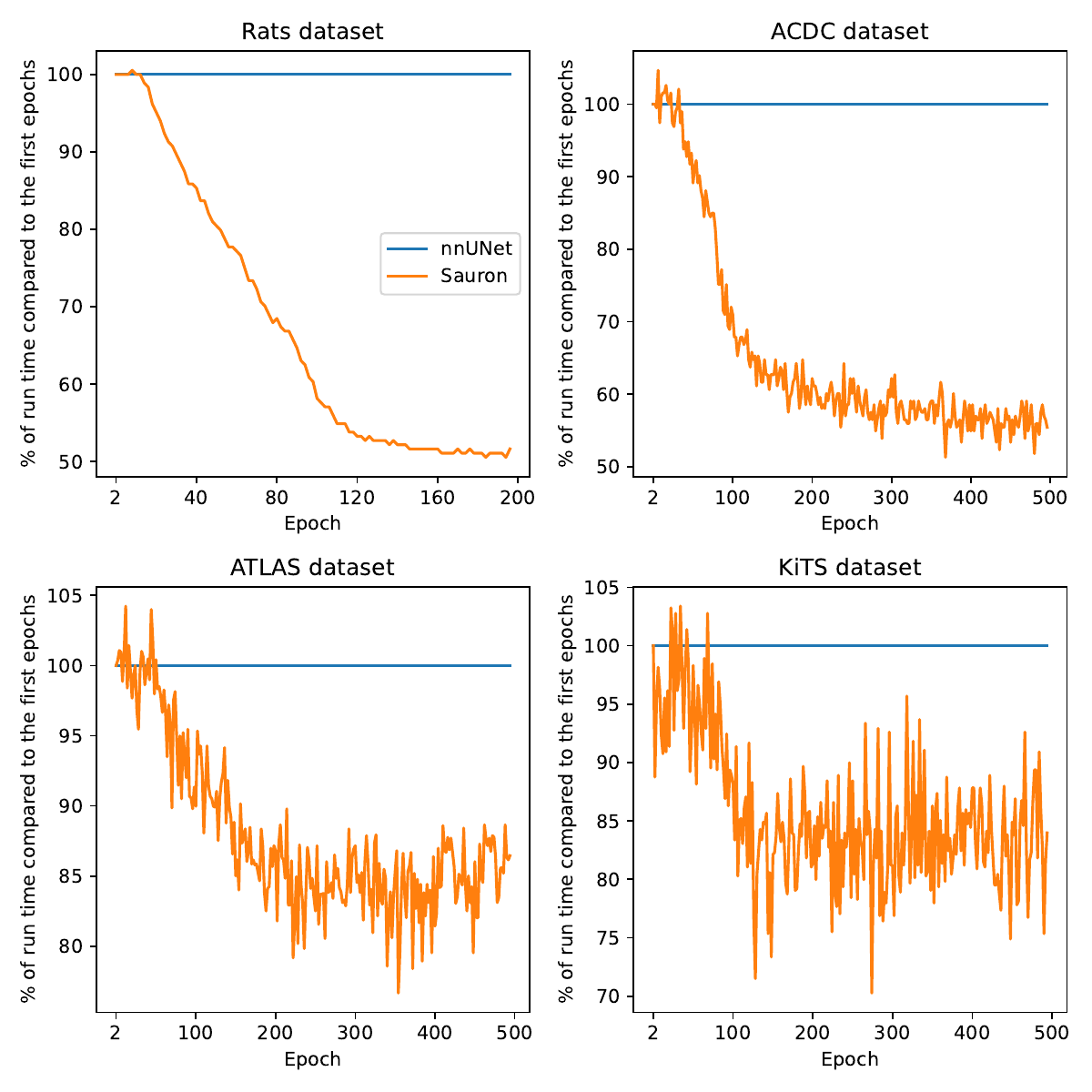} 
   \caption{Time required by each epoch divided by the time required by the first epoch during the optimization of nnUNet while pruning with Sauron.} \label{fig:asddd}
\end{figure}

\clearpage
\newpage

\section{nnUNet diagram} \label{appendix:nnunetdiagram}

\begin{figure}[!htb]
\centering
    \includegraphics[width=\textwidth]{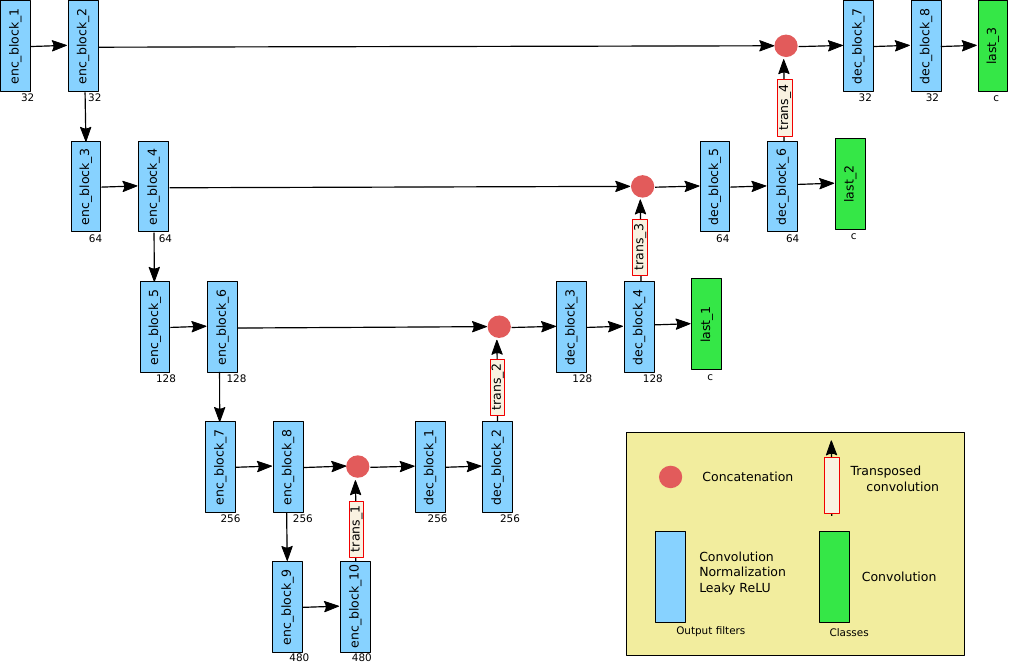} 
   \caption{Diagram of an archetypal nnUNet with five levels.} \label{fig:nnunet}
\end{figure}

nnUNet is a self-configurable U-Net optimized with extensive data augmentation, deep supervision, and polynomial learning rate decay.
In our experiments, the configuration of its architecture and optimization settings depended on the dataset, as in the original publication \citep{isensee2021nnu}.
The architectural components that depended on the dataset were the following:
\begin{itemize}
    \item \textbf{Number of levels}: Number of block pairs in the encoder with different feature map sizes. The number of levels in \cref{fig:nnunet} is five. After each even block in the decoder (except in \textit{dec\_block\_2}), nnUNet computes predictions at different resolutions, enabling deep supervision (green blocks in \cref{fig:nnunet}).
    \item \textbf{Number of filters}: Number of filters of the first two blocks in the encoder. The number of filters in every level doubles with respect to the previous level, unless it exceeds 480---maximum number of filters.
    \item \textbf{Normalization}: We employed either Batch normalization \citep{ioffe2015batch} or Instance normalization \citep{ulyanov2016instance}.
    \item \textbf{Dimensions}: Whether we used 3D or 2D convolutions.
\end{itemize}

nnUNet architecture, its optimization, dataset preprocessing, and data augmentation strategy varied across datasets.
Such disparity in configuration aimed to tailor each model and training settings to resemble as much as possible to previous studies that reported state-of-the-art performance \citep{isensee2017automatic,bernard2018deep,isensee2019automated,heller2021state,valverde2020ratlesnetv2}.
\Cref{table:summary_rats,table:summary_acdc,table:summary_atlas,table:summary_kits} list the configuration employed for each dataset.
This configuration can also be seen in our publicly-available code.

{\textbf{C.1. Preprocessing}}

Rats and ATLAS datasets were not preprocessed.
ACDC and KiTS datasets were resampled to their median voxel resolution (Tables \ref{table:summary_acdc} and \ref{table:summary_kits} report the final voxel resolution in mm.).
In KiTS dataset, images from patients 15, 23, 37, 68, 125, 133 were discarded due to their faulty ground-truth segmentation.
Intensity values were clipped to $[-79, 304]$ and normalized by subtracting $101$ and dividing by $76.9$.
Finally, images smaller than the patch size $160 \times 160 \times 80$ were padded.

{\textbf{C.2. Data augmentation}}

During training, images from Rats, ACDC, ATLAS, and KiTS datasets were augmented via TorchIO \citep{perez-garcia_torchio_2021}.
Images were randomly scaled and rotated with certain probability $p$.
Their intensity values were altered via random gamma correction.
Then, they were randomly flipped, and they were transformed via random elastic deformation.
Particularly in ACDC dataset, 2D slices from the 3D volumes were cropped or padded to $320 \times 320$ voxels.

{\textbf{C.3. Architecture}}

The number of levels of the nnUNet models trained on Rats, ATLAS, and KiTS datasets were five whereas in ACDC was seven.
nnUNet was optimized on Rats, ACDC, ATLAS, and KiTS datasets with 32, 48, 24, and 24 number of initial filters (\textit{enc\_block\_1}, \ref{appendix:nnunetdiagram}), respectively.
The nnUNet models optimized on Rats and ACDC datasets were 2D whereas the models for ATLAS and KiTS dataset were 3D.
Finally, the normalization layer utilized in Rats, ATLAS, and KiTS datasets was Instance Normalization \citep{ulyanov2016instance} whereas in ACDC was Batch Normalization \citep{ioffe2015batch}.

{\textbf{C.4. Optimization}}

All models were optimized with Adam \citep{kingma2014adam} with a starting learning rate of $10^{-3}$, weight decay of $10^{-5}$, and polynomial learning rate decay: $(1-(e/epochs))^{0.9}$.
For cSGD \citep{ding2019centripetal}, we used their proposed optimization strategy that is central to their approach.
nnUNet was optimized for 200 epochs in Rats dataset and 500 epochs in ACDC, ATLAS, and KiTS datasets.
The batch size in Rats, ACDC, ATLAS, and KiTS datasets was four, ten, one, and two, respectively.

\begin{table}[!htb]
  \caption{Rats dataset configuration summary}
  \label{table:summary_rats}
  \centering
  \begin{tabular}{llll}
    \toprule
    Preprocessing & Data augmentation & Architecture & Optimization \\
    \midrule
    —  & Scale $[0.9, 1.1], p=0.5$ & Five levels & 200 Epochs     \\
    & Rotation $[-10, 10], p=0.5$ & 32 Init. filters & Batch size: 4     \\
    & Gamma correction $[-0.3, 0.3], p=0.5$ & 2D &      \\
    & Flip axis $p=0.5$ & Instance Norm. &      \\
    \bottomrule
  \end{tabular}
\end{table}

\begin{table}[!htb]
  \caption{ACDC dataset configuration summary}
  \label{table:summary_acdc}
  \centering
  \begin{tabular}{llll}
    \toprule
    Preprocessing & Data augmentation & Architecture & Optimization \\
    \midrule
    Resample (1.25, 125, 1) & Scale $[0.85, 1.25], p=0.2$ & Seven levels & 500 Epochs      \\
    & Rotation $[-180, 180], p=0.2$ & 48 Init. filters & Batch size: 10\\
    & Elastic deformation $p=0.3$ & 2D &       \\
    & Gamma correction $[-0.3, 0.5], p=0.3$ & Batch Norm. &      \\
    & Flip axis $p=0.5$ &  &       \\
    & CropOrPad (320, 320) &  &       \\
    \bottomrule
  \end{tabular}
\end{table}

\begin{table}[!htb]
  \caption{ATLAS dataset configuration summary}
  \label{table:summary_atlas}
  \centering
  \begin{tabular}{llll}
    \toprule
    Preprocessing & Data augmentation & Architecture & Optimization \\
    \midrule
    — & Scale $[0.85, 1.25], p=0.2$ & Five levels & 500 Epochs      \\
    & Rotation $[-180, 180], p=0.2$ & 24 Init. filters & Batch size: 1\\
    & Elastic deformation $p=0.3$ & 3D &       \\
    & Gamma correction $[-0.3, 0.5], p=0.3$ & Instance Norm. &      \\
    & Flip axis $p=0.5$ &  &       \\
    \bottomrule
  \end{tabular}
\end{table}

\begin{table}[!htb]
  \caption{KiTS dataset configuration summary}
  \label{table:summary_kits}
  \centering
  \begin{tabular}{llll}
    \toprule
    Preprocessing & Data augmentation & Architecture & Optimization \\
    \midrule
    Discard samples  & Scale $[0.85, 1.25], p=0.2$ & Five levels & 500 Epochs  \\
    Resample (1.62, 1.62, 3.22)  & Rotation $[-180, 180], p=0.2$ & 24 Init. filters & Batch size: 2  \\
    Clip intensities (-79, 304)  & Elastic deformation $p=0.3$ & 3D &   \\
    Norm. $(data - 101) / 76.9$  & Gamma correction $[-0.3, 0.5], p=0.3$ & Instance Norm. &   \\
    Pad $(160 \times 160 \times 80)$  & Flip axis $p=0.5$ &  &   \\
    \bottomrule
  \end{tabular}
\end{table}

\clearpage
\newpage

\section{Segmentation results}
\begin{figure}[!htb]
\centering
    \includegraphics[width=\textwidth]{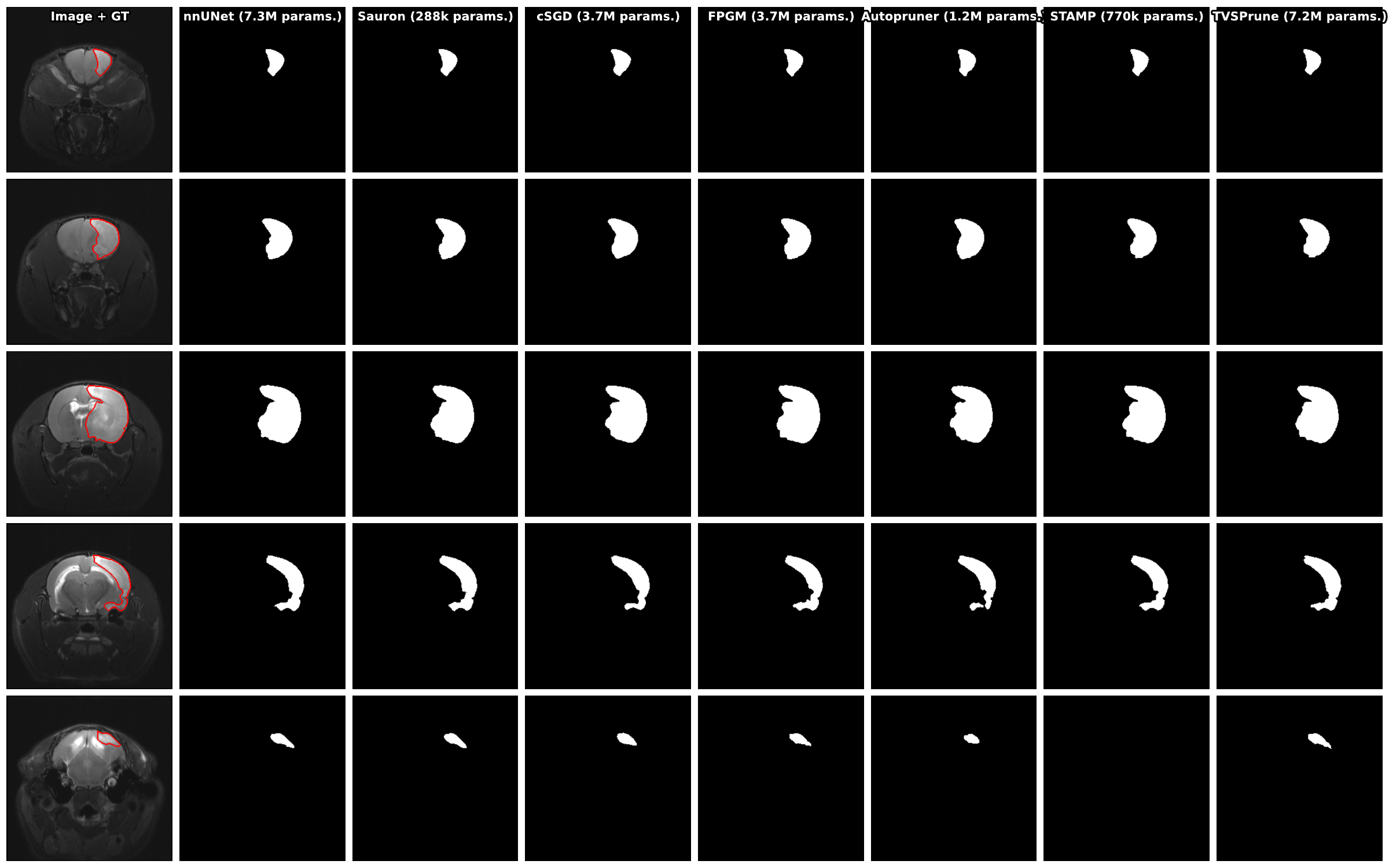} 
   \caption{Example segmentation in Rats dataset. First column: image and ground truth. Second column: nnUNet segmentation (baseline). Other columns: segmentations from nnUNet pruned by the compared methods. Parenthesis: number of parameters of nnUNet after pruning.} \label{fig:r1}
\end{figure}

\begin{figure}[!htb]
\centering
    \includegraphics[width=\textwidth]{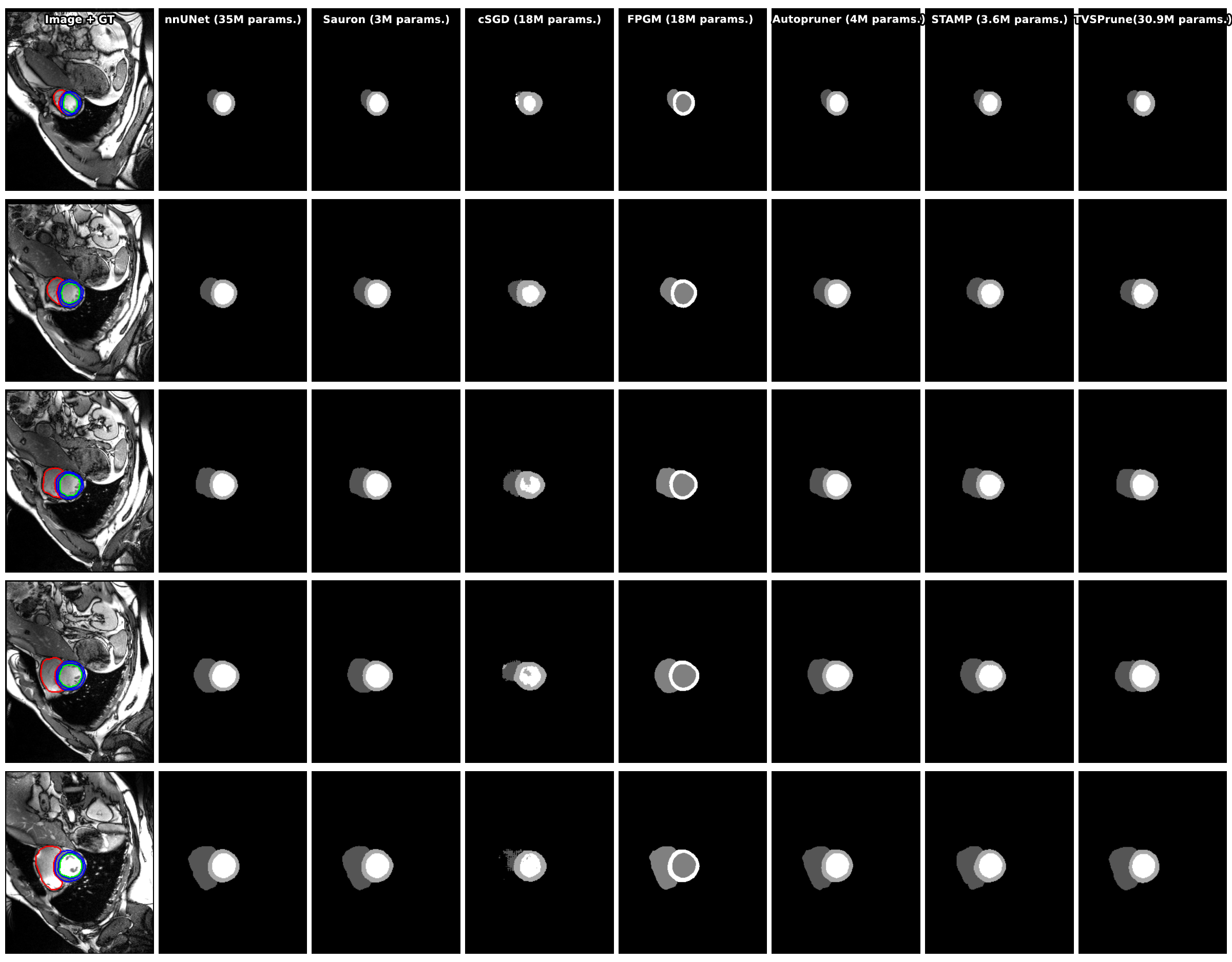} 
   \caption{Example segmentation in ACDC dataset. First column: image and ground truth. Second column: nnUNet segmentation (baseline). Other columns: segmentations from nnUNet pruned by the compared methods. Parenthesis: number of parameters of nnUNet after pruning.} \label{fig:r2}
\end{figure}

\begin{figure}[!htb]
\centering
    \includegraphics[width=\textwidth]{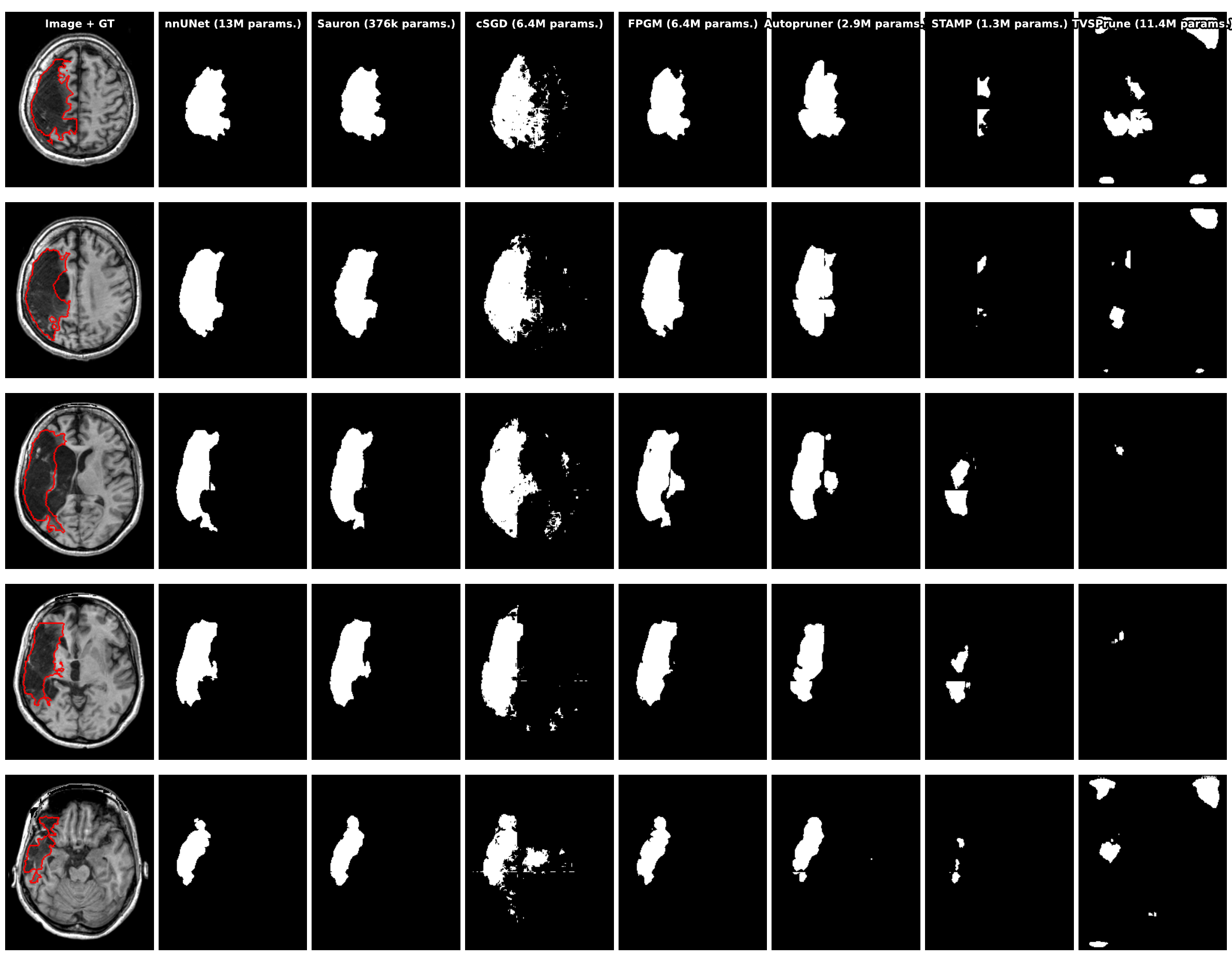} 
   \caption{Example segmentation in ATLAS dataset. First column: image and ground truth. Second column: nnUNet segmentation (baseline). Other columns: segmentations from nnUNet pruned by the compared methods. Parenthesis: number of parameters of nnUNet after pruning.} \label{fig:r3}
\end{figure}

\begin{figure}[!htb]
\centering
    \includegraphics[width=\textwidth]{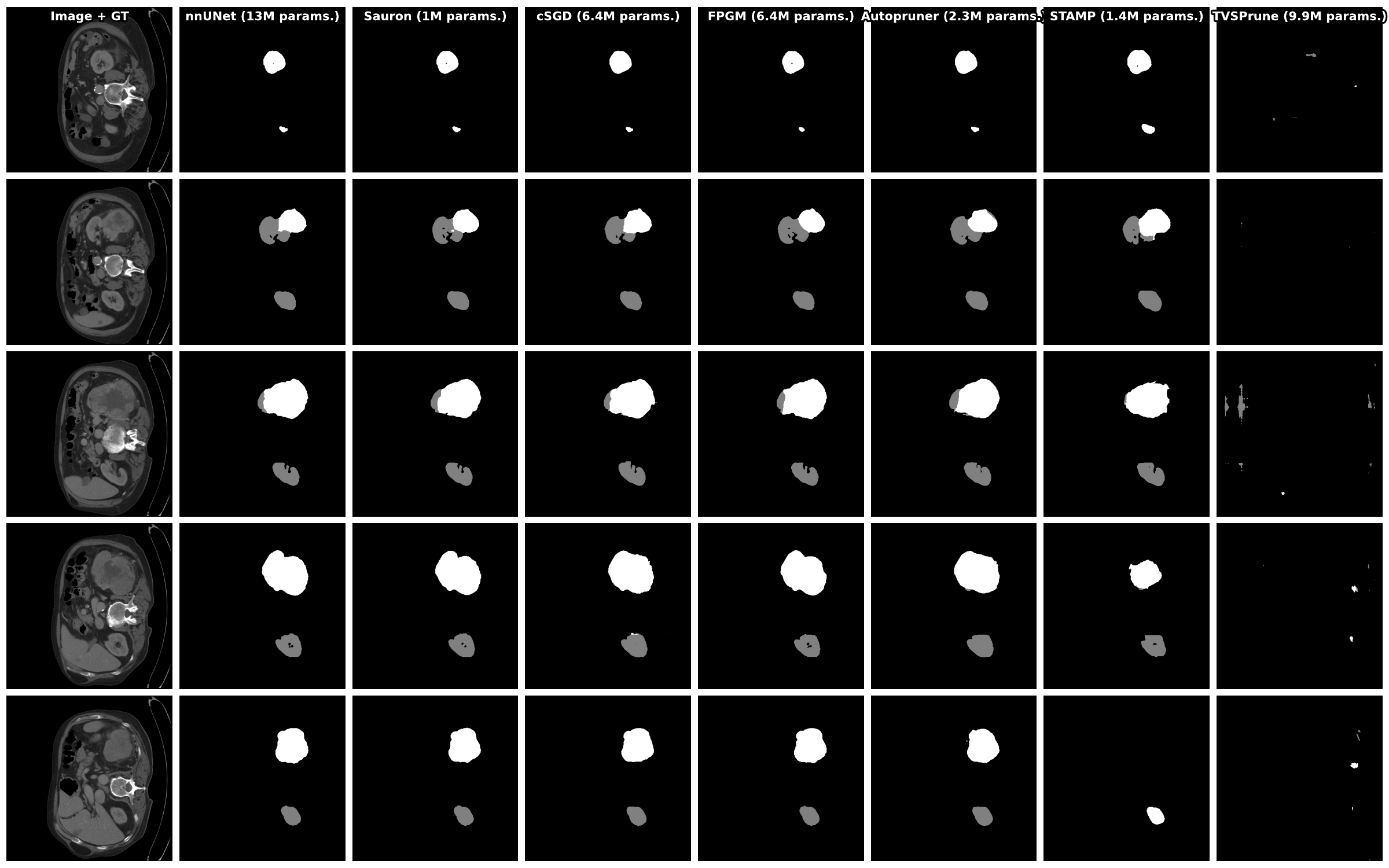} 
   \caption{Example segmentation in KiTS dataset. First column: image and ground truth. Second column: nnUNet segmentation (baseline). Other columns: segmentations from nnUNet pruned by the compared methods. Parenthesis: number of parameters of nnUNet after pruning.} \label{fig:r4}
\end{figure}

\clearpage
\newpage

\section{Sauron's training and validation loss}

\begin{figure}[!htb]
\centering
    \includegraphics[width=\textwidth]{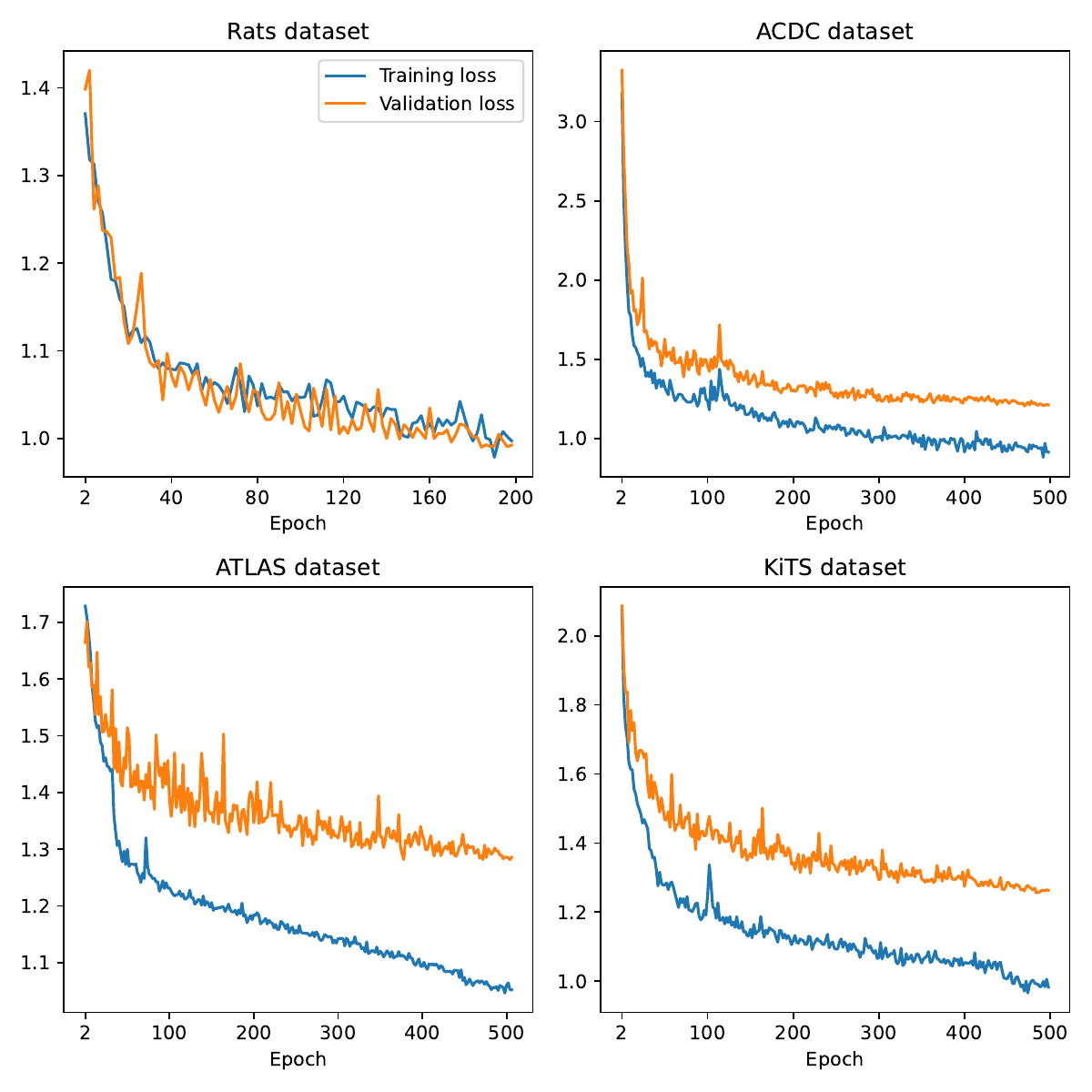} 
   \caption{Training and validation loss when optimizing nnUNet while pruning with Sauron.} \label{fig:ggg}
\end{figure}

\clearpage
\newpage
\section{Increase/decrease in clusterability metrics}
\Cref{table:clusterability_metrics} lists the relative increase/decrease in the three clusterability measures (dip-test value, distances $\delta_{opt}$, and average number of neighbors) for each convolutional layer.
The increase/decrease is computed as the ratio between $p_1$ and $p_2$, where $p_1$ is the average value during the first third of the optimization, and $p_2$ is the average value in the last third of the training.
An increase in clusterability is indicated by 1) an increase in dip-test, 2) a decrease in $\delta_{opt}$, and 3) an increase in the average number of neighbors.

\begin{table}[!htb]
  \caption{Name of the convolutional layer (see \Cref{appendix:nnunetdiagram}), number of output filters, and relative increase/decrease in three clusterability measures. Gray: layers with 256 or more feature maps.}
  \label{table:clusterability_metrics}
  \centering
  \begin{tabular}{lllll}
    \toprule
    Conv. Layer & Filters & Dip-test & Distances $\delta_{opt}$ & Avg. neighbors \\
    \midrule
enc\_conv\_1 & 32 & -46.0\% & 2.1\% & -74.3\% \\
enc\_conv\_2 & 32 & -13.5\% & 5.4\% & -91.5\% \\
enc\_conv\_3 & 64 & -15.0\% & 2.4\% & -97.9\% \\
enc\_conv\_4 & 64 & -4.2\% & -13.7\% & -99.8\% \\
enc\_conv\_5 & 128 & -5.0\% & -7.4\% & -93.6\% \\
enc\_conv\_6 & 128 & 3.3\% & -9.4\% & -97.5\% \\
\rowcolor{Gray} enc\_conv\_7 & 256 & 688.0\% & -22.0\% & 75.1\% \\
\rowcolor{Gray} enc\_conv\_8 & 256 & 168.5\% & -29.3\% & 118.5\% \\
\rowcolor{Gray} enc\_conv\_9 & 480 & -32.4\% & -53.4\% & 128.9\% \\
\rowcolor{Gray} enc\_conv\_10 & 480 & 118.1\% & -51.1\% & 161.4\% \\
\rowcolor{Gray} dec\_trans\_1 & 256 & 226.7\% & -29.7\% & 30.7\% \\
\rowcolor{Gray} dec\_conv\_1 & 256 & 212.7\% & -42.4\% & 63.8\% \\
\rowcolor{Gray} dec\_conv\_2 & 256 & 121.9\% & -8.3\% & 70.3\% \\
dec\_trans\_2 & 128 & 18.0\% & -55.0\% & 30.3\% \\
dec\_conv\_3 & 128 & 89.0\% & 3.4\% & 3.4\% \\
dec\_conv\_4 & 128 & -15.2\% & 13.7\% & 192.5\% \\
dec\_trans\_3 & 64 & 214.6\% & -40.8\% & 1872.5\% \\
dec\_conv\_5 & 64 & -7.4\% & 0.0\% & 49.1\% \\
dec\_conv\_6 & 64 & 22.9\% & -3.1\% & 312.9\% \\
dec\_trans\_4 & 32 & 130.8\% & -47.1\% & 2769.6\% \\
dec\_conv\_7 & 32 & -9.2\% & 5.5\% & 0\% \\
dec\_conv\_8 & 32 & 7.6\% & 9.0\% & -97.4\% \\

    \bottomrule
  \end{tabular}
\end{table}

\clearpage
\newpage
\section{Feature maps in the second-to-last convolutional layer of the baseline (unpruned) nnUNet models}

\begin{figure}[!htb]
\centering
    \includegraphics[width=\textwidth]{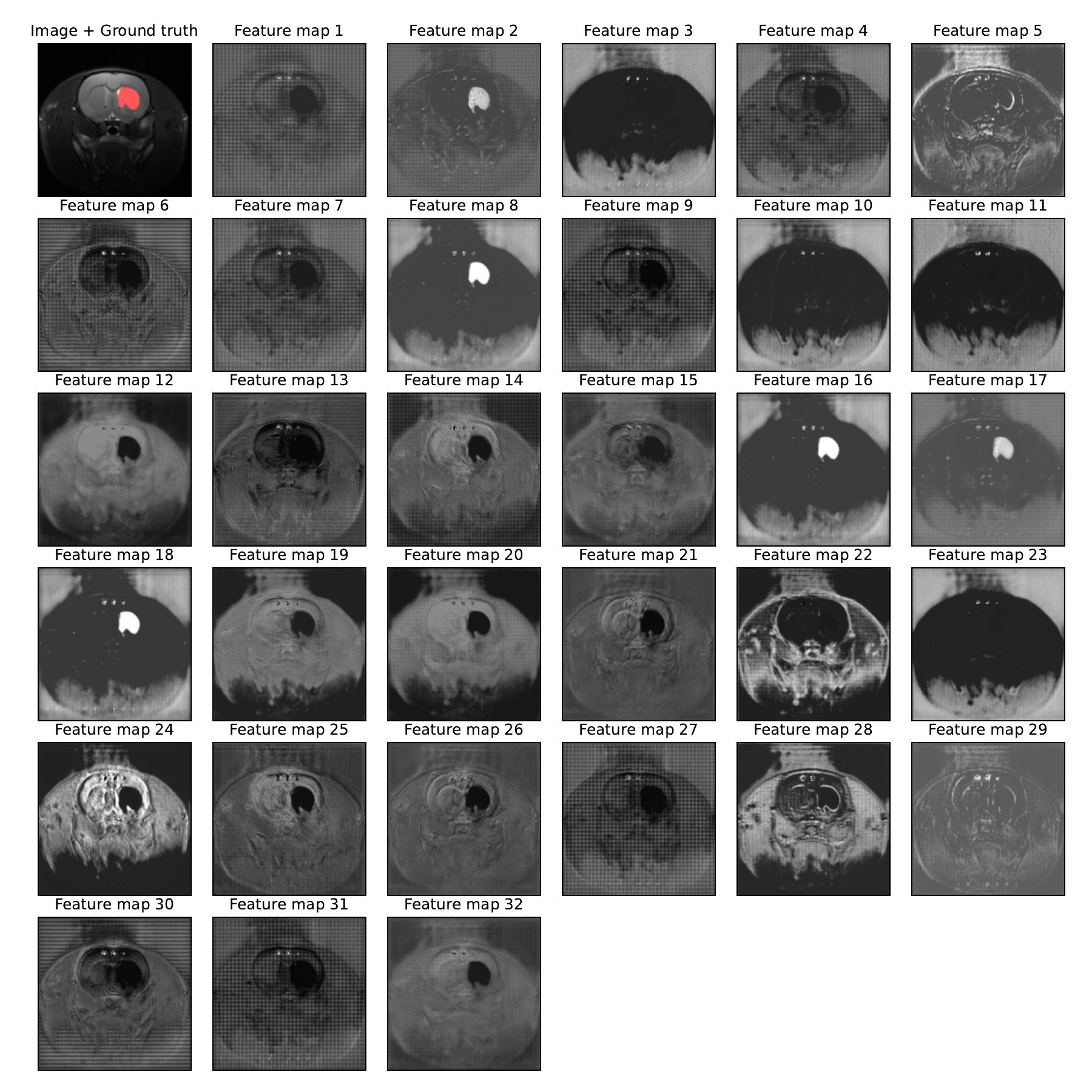} 
   \caption{Rats dataset.} \label{fig:inter_rats}
\end{figure}

\begin{figure}[!htb]
\centering
    \includegraphics[width=\textwidth]{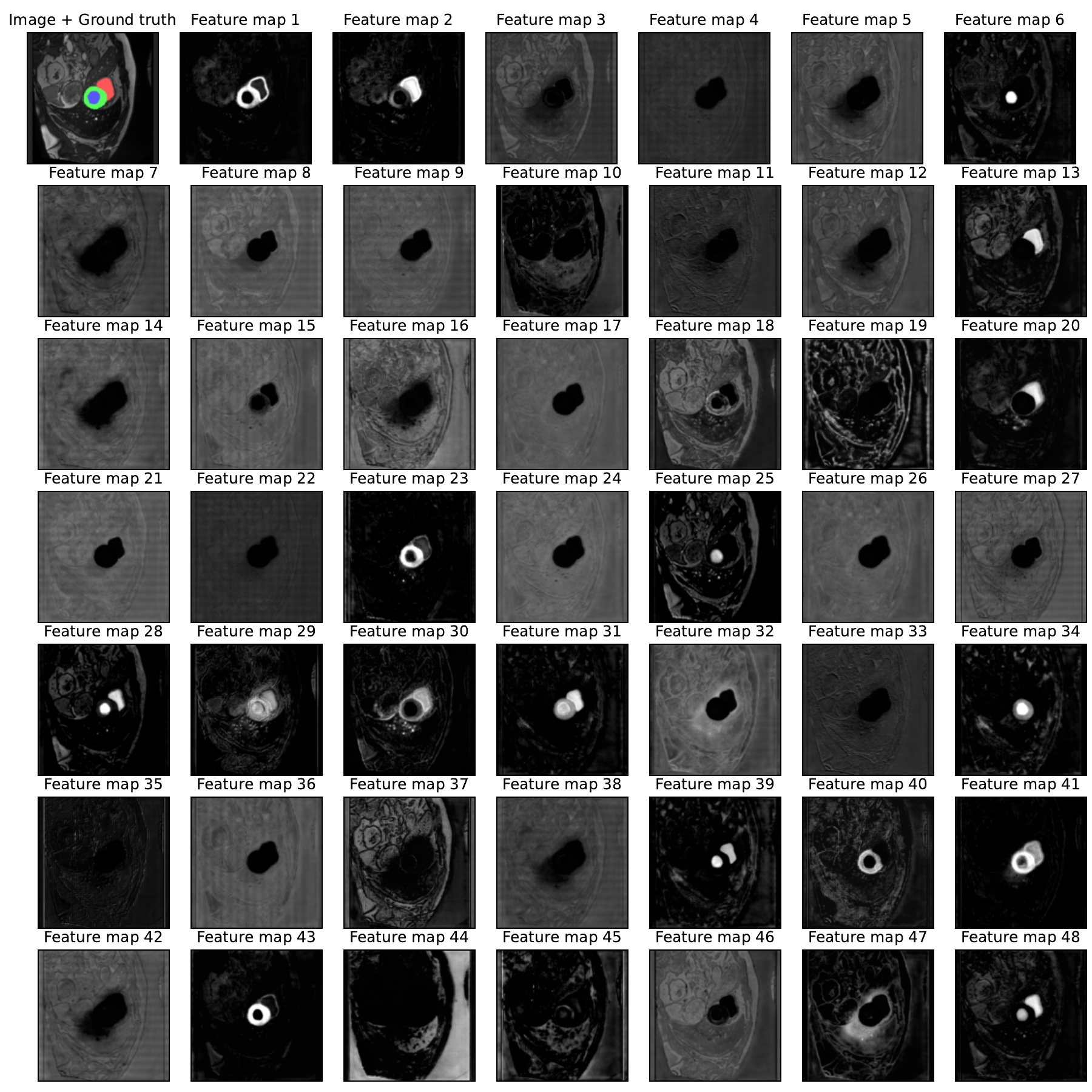} 
   \caption{ACDC dataset.} \label{fig:inter_acdc}
\end{figure}

\begin{figure}[!htb]
\centering
    \includegraphics[width=\textwidth]{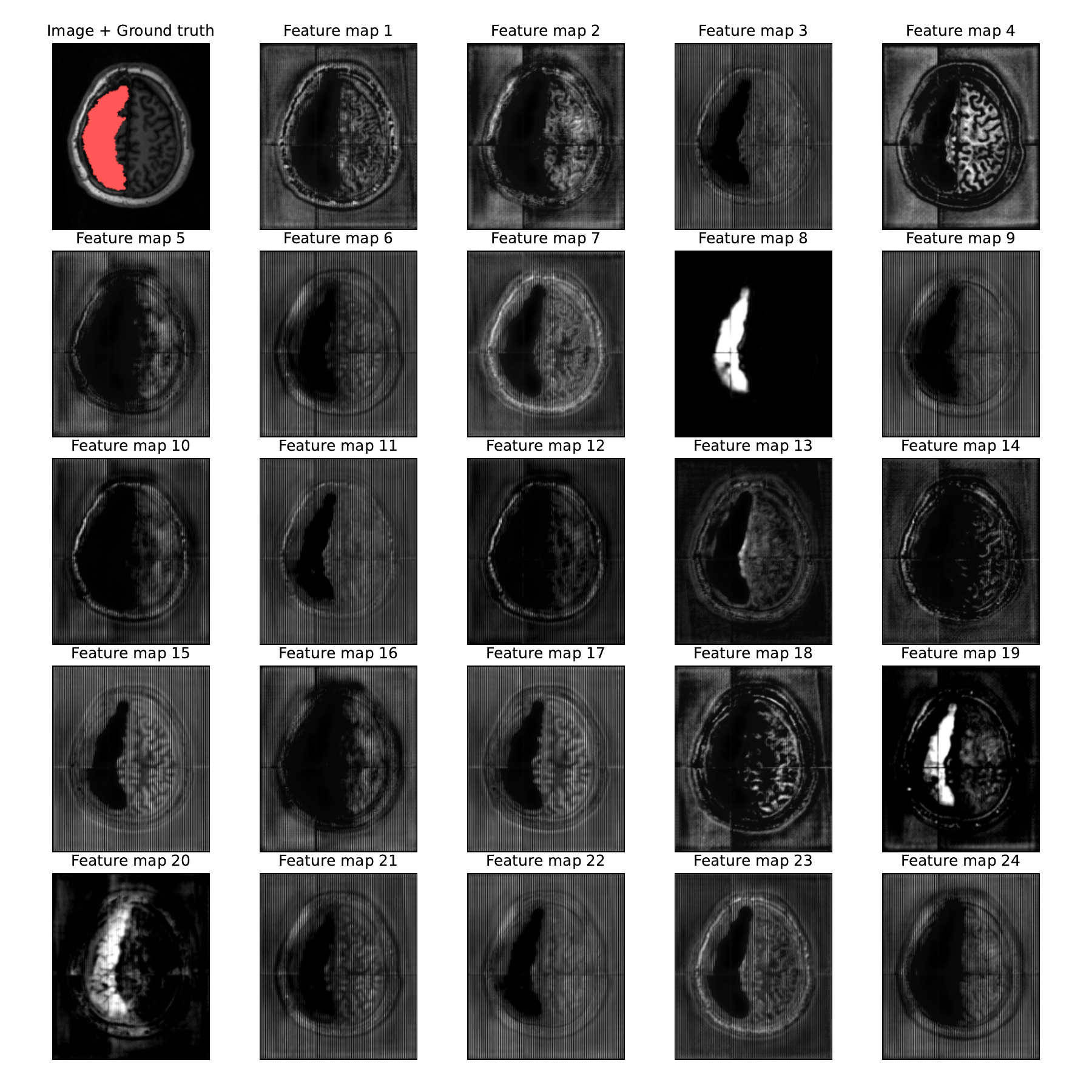} 
   \caption{ATLAS dataset.} \label{fig:inter_atlas}
\end{figure}

\begin{figure}[!htb]
\centering
    \includegraphics[width=\textwidth]{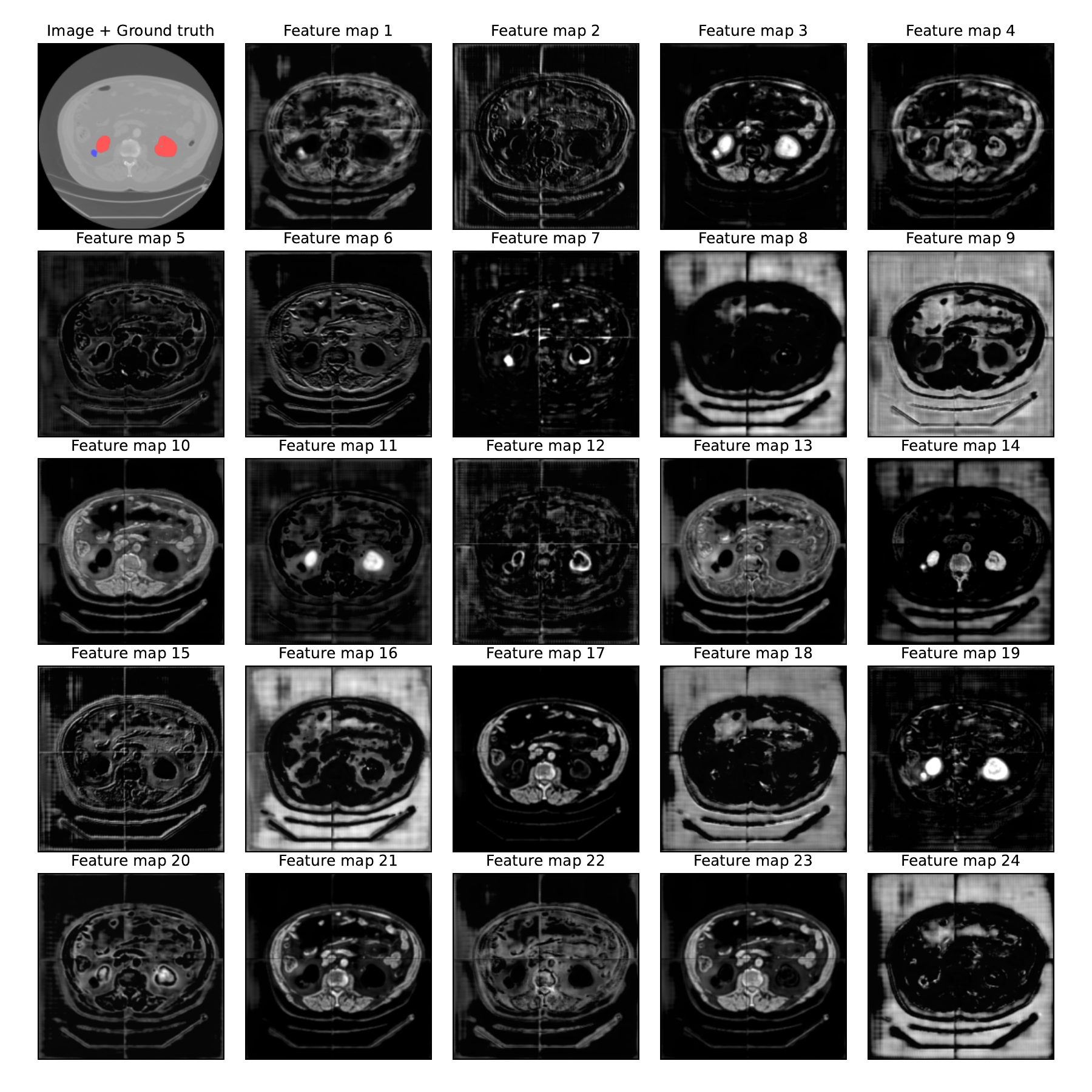} 
   \caption{KiTS dataset.} \label{fig:inter_kits}
\end{figure}

\end{document}